\documentclass[letterpaper]{article} % DO NOT CHANGE THIS
\usepackage{aaai24}  % DO NOT CHANGE THIS
\usepackage{times}  % DO NOT CHANGE THIS
\usepackage{helvet}  % DO NOT CHANGE THIS
\usepackage{courier}  % DO NOT CHANGE THIS
\usepackage[hyphens]{url}  % DO NOT CHANGE THIS
\usepackage{graphicx} % DO NOT CHANGE THIS
\usepackage{natbib}  % DO NOT CHANGE THIS AND DO NOT ADD ANY OPTIONS TO IT
\usepackage{caption} % DO NOT CHANGE THIS AND DO NOT ADD ANY OPTIONS TO IT
\usepackage{algorithm}
\usepackage{algorithmic}

\usepackage{multirow}
\usepackage{subfig}
\usepackage{float}
\usepackage{multicol}

\usepackage{amsmath}
\usepackage{amsthm}
\usepackage{booktabs}
\usepackage{algorithm}
\usepackage{algorithmic}
\usepackage[switch]{lineno}
\usepackage{multirow}
\usepackage{graphicx}
\usepackage{subfig}
\usepackage{xcolor}

\usepackage{amsfonts}       % blackboard math symbols
\usepackage{nicefrac}       % compact symbols for 1/2, etc.
\usepackage{microtype}      % microtypography
\usepackage{xcolor}         % colors
\usepackage{graphicx}
\usepackage[most]{tcolorbox}
\usepackage{multirow}    
\usepackage{amsmath,amssymb,amsfonts}
\usepackage{subfig}
\usepackage{pifont}

\usepackage{tabularx}
\usepackage{ltablex}
\newcolumntype{b}{X}
\newcolumntype{s}{>{\hsize=2.2\hsize}X}

\usepackage{newfloat}
\usepackage{listings}
\DeclareCaptionStyle{ruled}{labelfont=normalfont,labelsep=colon,strut=off} % DO NOT CHANGE THIS
\floatstyle{ruled}
\newfloat{listing}{tb}{lst}{}
\floatname{listing}{Listing}
\title{Preliminary Investigations of a Multi-Faceted Robust and Synergistic Approach in Semiconductor Electron Micrograph Analysis: Integrating Vision Transformers with Large Language and Multimodal Models}
\author{
    Sakhinana Sagar Srinivas\textsuperscript{\rm 1}\thanks{Designed, programmed the software, and drafted manuscript.},
    \{Geethan Sannidhi\textsuperscript{\rm 2}, Sreeja Gangasani\textsuperscript{\rm 3}, Chidaksh Ravuru\textsuperscript{\rm 4}\}\thanks{Conducted experiments and analyzed visual results}, 
    Venkataramana Runkana\textsuperscript{\rm 1}\\
}
\affiliations{
    \textsuperscript{\rm 1}TCS Research,
    \textsuperscript{\rm 2}IIIT Pune,
    \textsuperscript{\rm 3}IIT Palakkad,
    \textsuperscript{\rm 4}IIT Dharwad \\
    \texttt{sagar.sakhinana@tcs.com}, \texttt{geethan.iiitp.ac.in}, \texttt{111901023@smail.iitpkd.ac.in}, \texttt{200010046@iitdh.ac.in}, \texttt{venkat.runkana@tcs.com}\\
}

\usepackage{bibentry}
\begin{document}

\maketitle

\vspace{-2mm}
\begin{abstract}
\vspace{-2mm}
Characterizing materials using electron micrographs is crucial in areas such as semiconductors and quantum materials. Traditional classification methods falter due to the intricate structures of these micrographs. This study introduces an innovative architecture that leverages the generative capabilities of zero-shot prompting in Large Language Models (LLMs) such as GPT-4(language only), the predictive ability of few-shot (in-context) learning in Large Multimodal Models (LMMs) such as GPT-4(V)ision, and fuses knowledge across image-based and linguistic insights for accurate nanomaterial category prediction. This comprehensive approach aims to provide a robust solution for the automated nanomaterial identification task in semiconductor manufacturing, blending performance, efficiency, and interpretability. Our method surpasses conventional approaches, offering precise nanomaterial identification and facilitating high-throughput screening.
\end{abstract}

\vspace{-6mm}
\section{Introduction}
\begin{figure}[htbp]
     \centering
     \subfloat[High intra-class dissimilarity: \textit{MEMS} devices exhibit a high degree of heterogeneity.]{\includegraphics[height=0.08\textwidth]{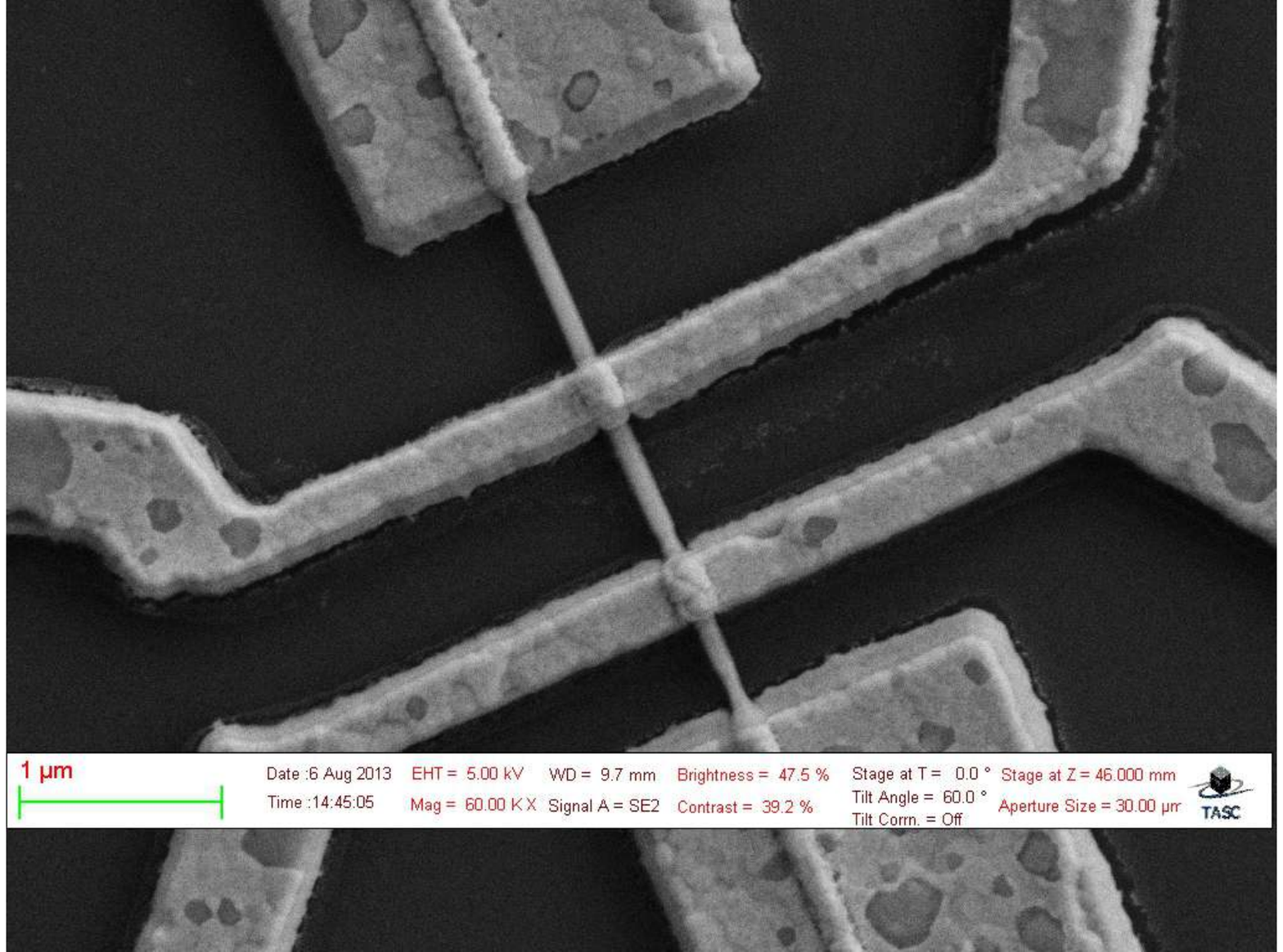}
     \includegraphics[height=0.08\textwidth]{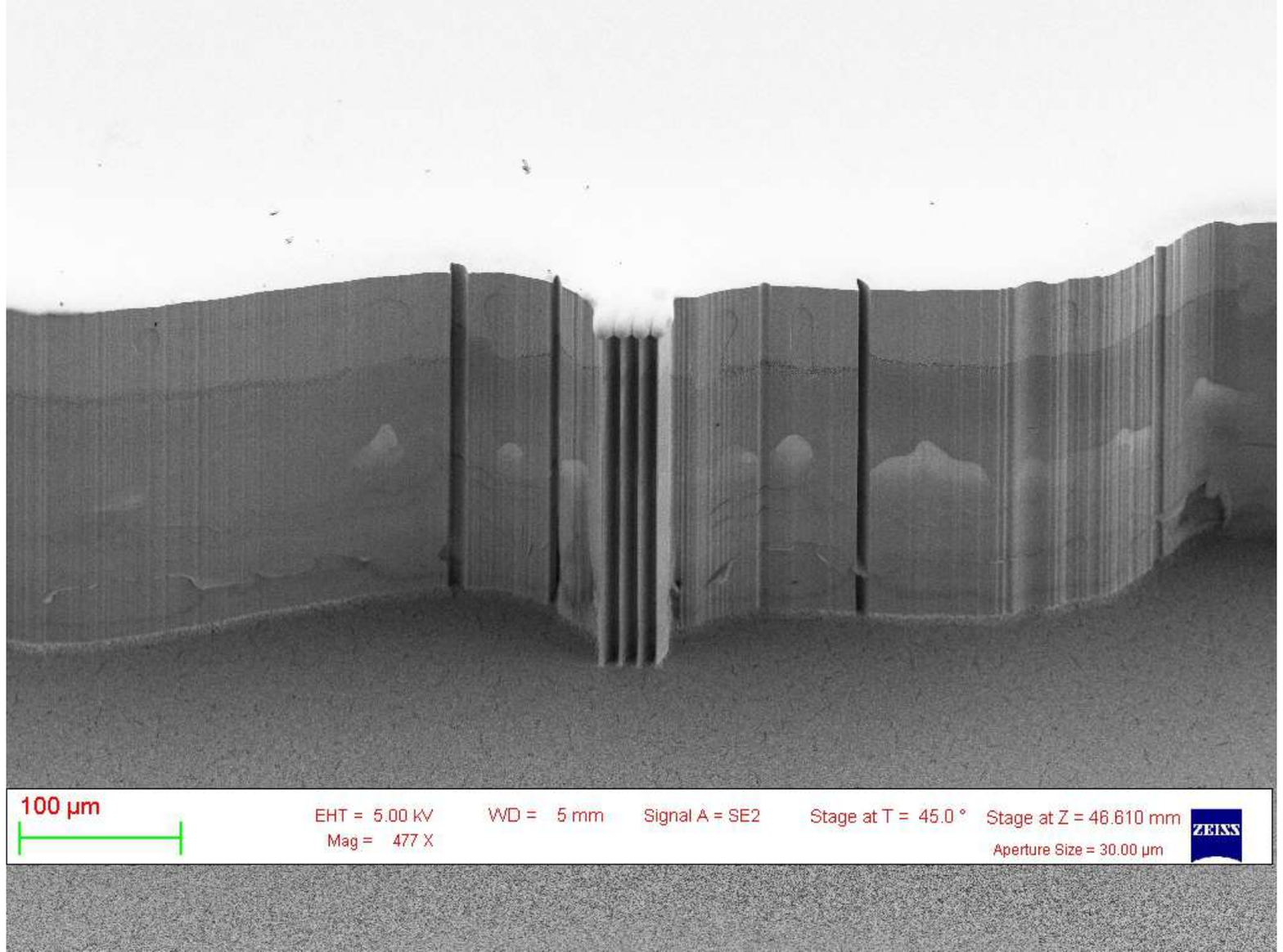}
     \includegraphics[height=0.08\textwidth]{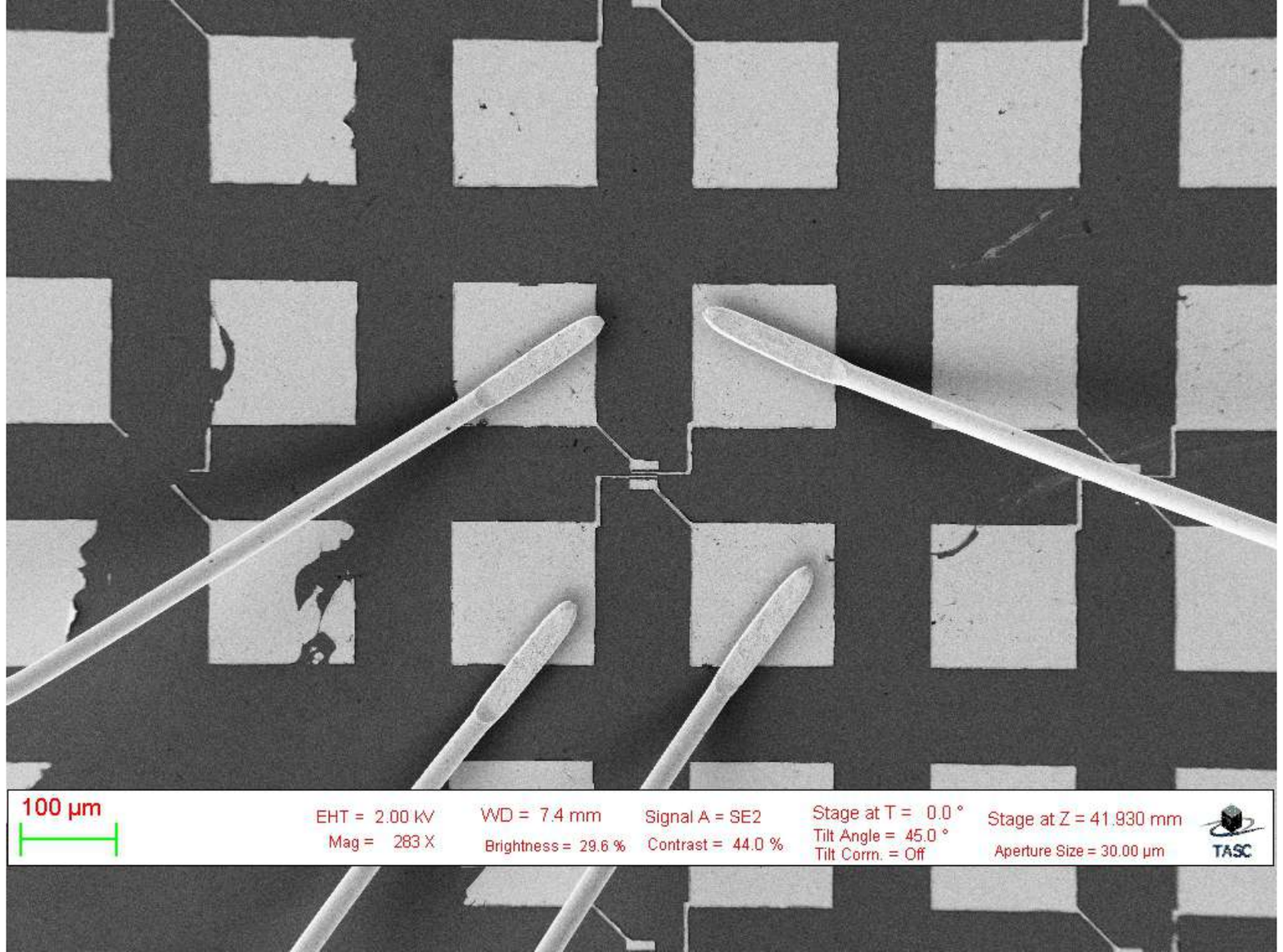}
     }
     \vspace{-0.5mm}
     \qquad
     \subfloat[High inter-class similarity: Different nanomaterial categories (\textit{listed from left to right as porous sponges, particles and powders}), are similar-looking or indistinguishable.]{\includegraphics[height=0.08\textwidth]{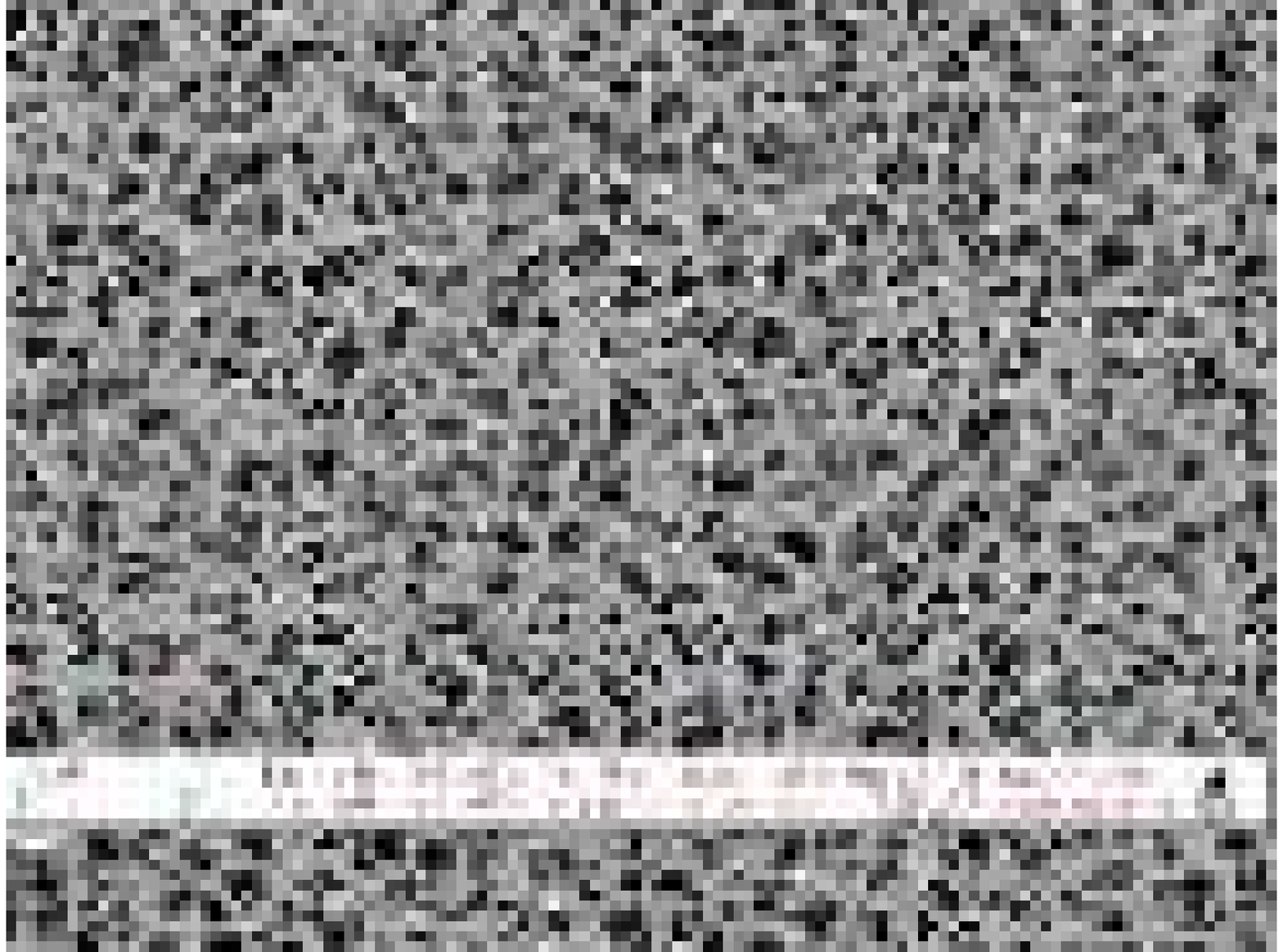}
     \includegraphics[height=0.08\textwidth]{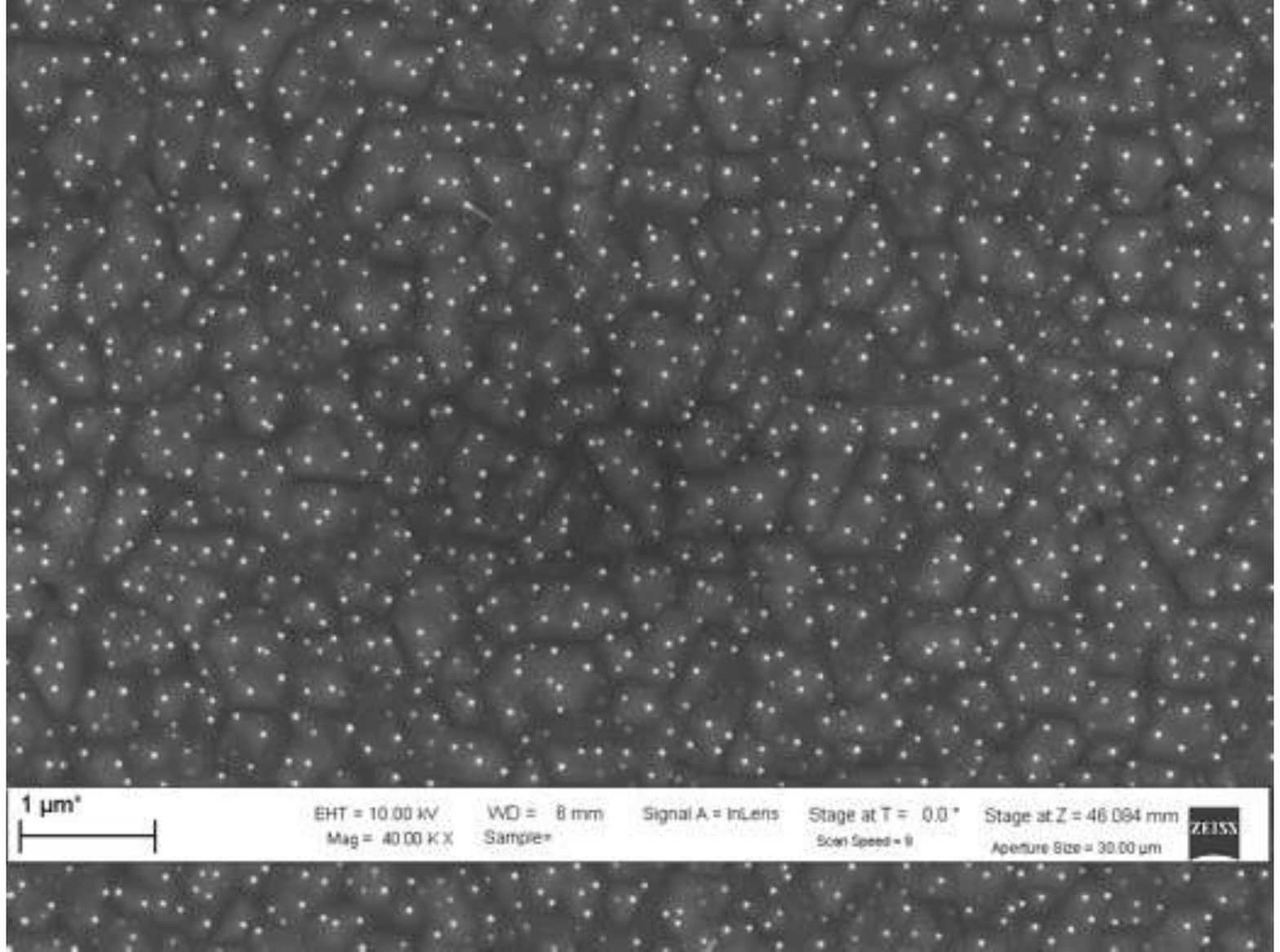}
     \includegraphics[height=0.08\textwidth]{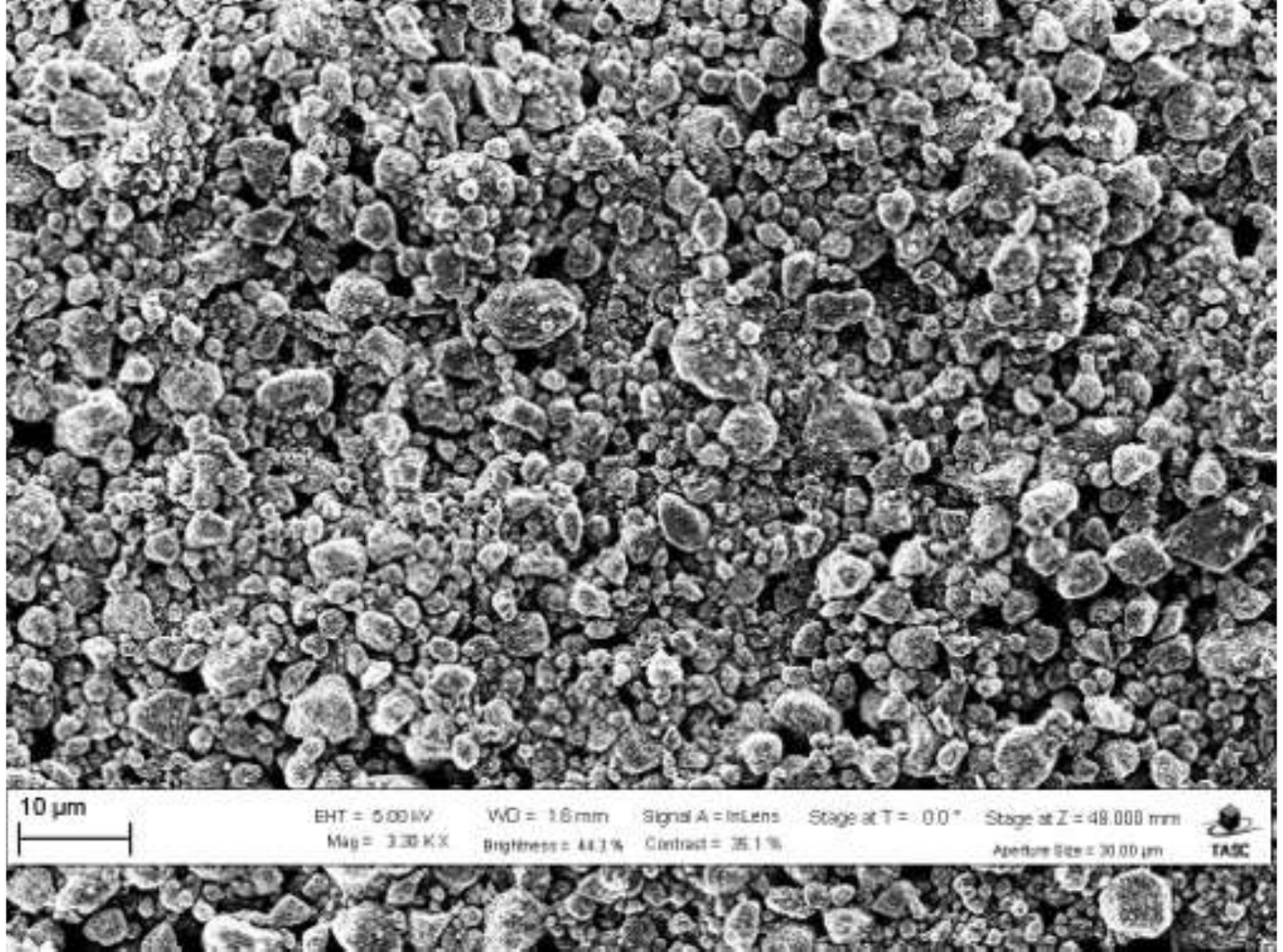}
     }
     \vspace{-0.5mm}
     \qquad
     \subfloat[Multi-spatial scale (spatial heterogeneity) patterns in electron micrographs of \textit{nanoparticles} are evident.]{\includegraphics[height=0.08\textwidth]{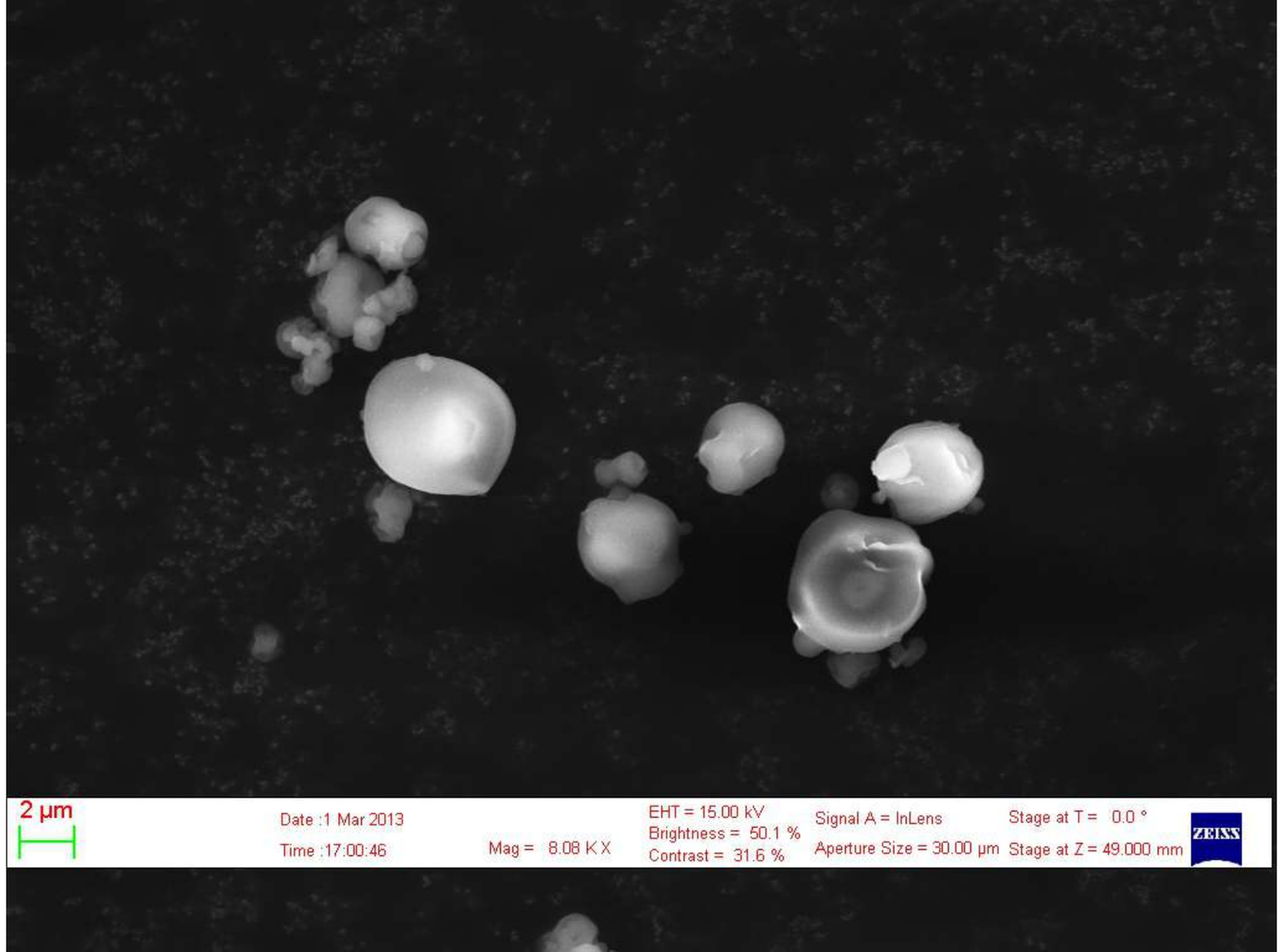}
     \includegraphics[height=0.08\textwidth]{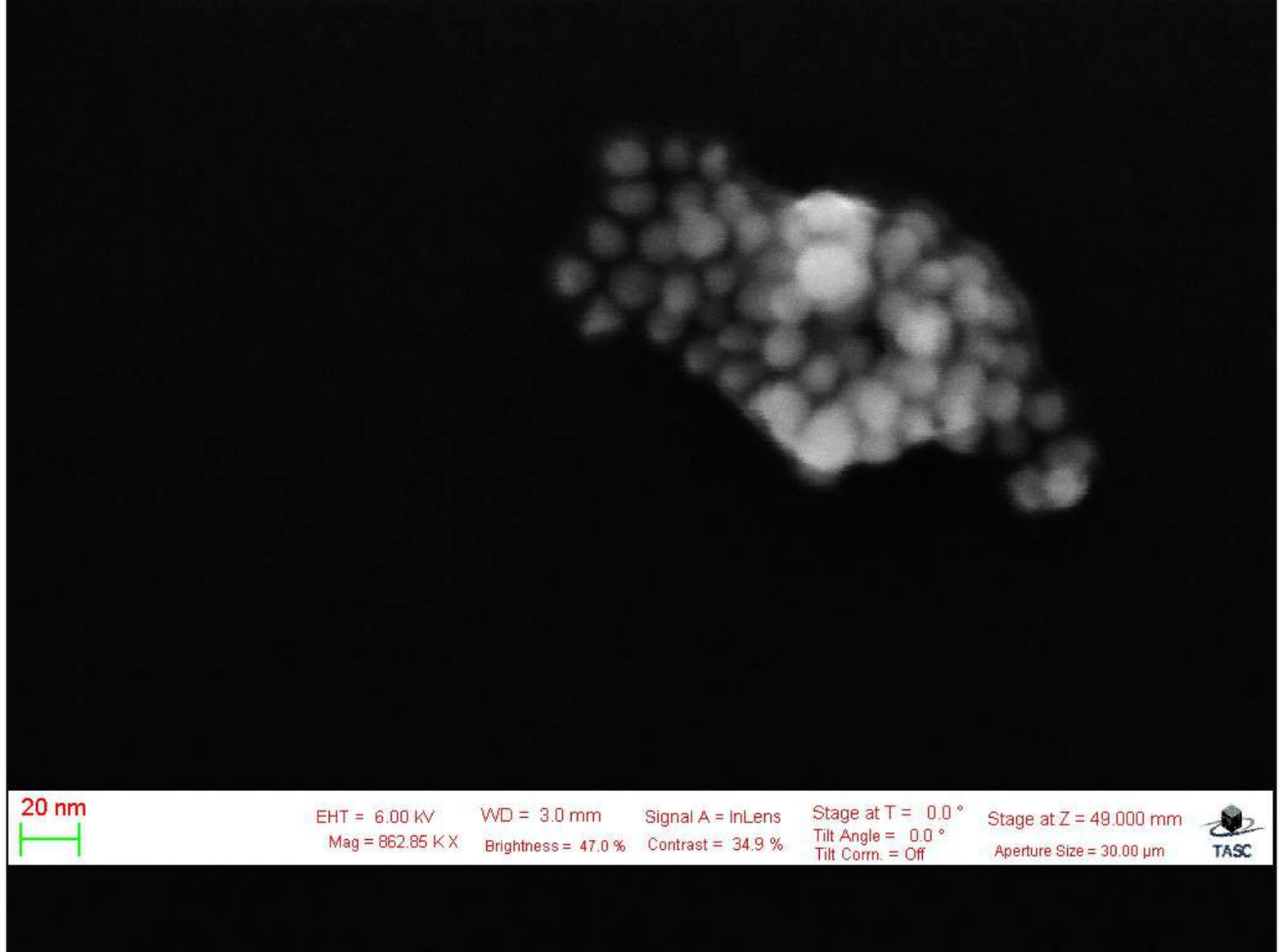}
     \includegraphics[height=0.08\textwidth]{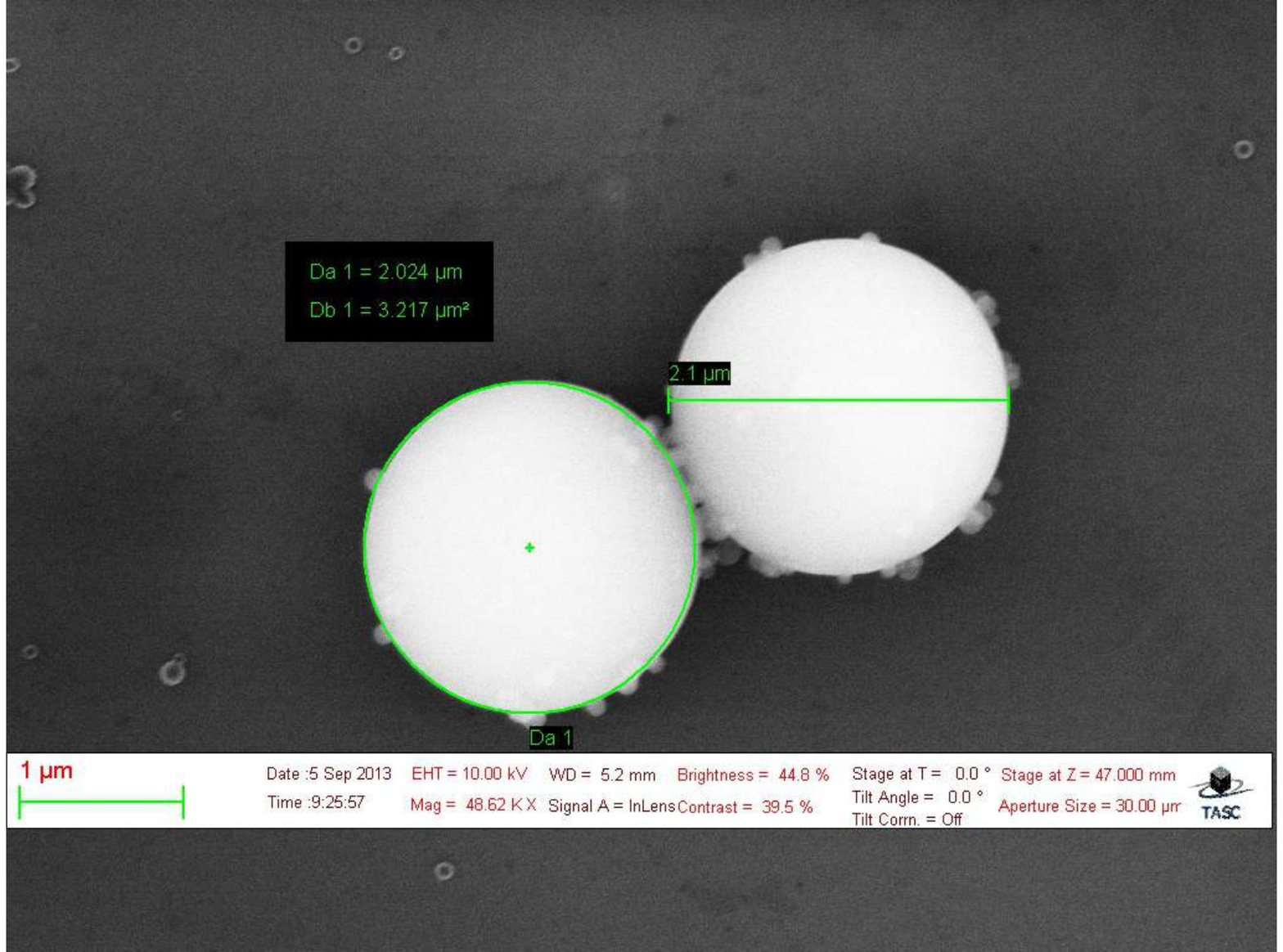}
     }
     \vspace{-3mm}
     \caption{The figure provides a visual representation of the challenges of classifying electron micrographs in the SEM dataset\cite{aversa2018first}.}
     \vspace{-4mm}
     \label{fig:figure1}
\end{figure}

\vspace{-1mm}
Semiconductors have been the backbone of technological advancements in modern electronics, driving growth and innovation in computing and communication systems, among others. The semiconductor process comprises three main stages: (a) design and development, during which fabless firms create chip blueprints, specifying the architecture, functions, and specifications of the miniaturized chips; (b) fabrication, where specialized foundries manufacture chips by etching integrated circuits onto silicon wafers using intricate technologies; and (c) testing and assembly, during which chips undergo rigorous testing and are subsequently assembled into protective packages for integration into electronic devices. This collective effort results in the production of high-quality semiconductor components suitable for a wide range of applications. The state-of-the-art imaging and analysis methods\cite{holt2013sem} are indispensable in semiconductor manufacturing for the development of next-generation miniaturized chips, especially those sized at 7 nm or smaller. The pursuit of miniaturized chips below 7 nm technologies introduces a level of complexity and precision that significantly increases the risk of errors in the manufacturing process. These errors can compromise the consistency of high-quality chips and amplify the variability in chip performance, posing a substantial challenge for manufacturers aiming to produce reliable and advanced chips at this scale. The semiconductor industry utilizes various advanced electron beam tools, including scanning and transmission electron microscopy, to create images or micrographs of semiconductor materials, structures, and devices at the micro and nanoscale with high resolution and detail. These tools contribute to quality control, process monitoring, failure analysis, and materials characterization in the semiconductor industry. Automated labeling of electron micrographs, though advantageous, poses a considerable challenge due to the level of detail, complexity of patterns, and information density involved. These challenges arise primarily from the high inter-category similarity (similar-looking or indistinguishable) between different nanomaterials, high intra-category dissimilarity within nanomaterials (distinct or differing appearances), and the presence of intricate visual patterns in nanomaterials across various scales (spatial heterogeneity). The complexities of automated nanomaterial identification tasks are illustrated in Figure \ref{fig:figure1}. 

\vspace{-4mm}

\begin{figure}[htbp]
\centering
\includegraphics[width=0.35\textwidth]{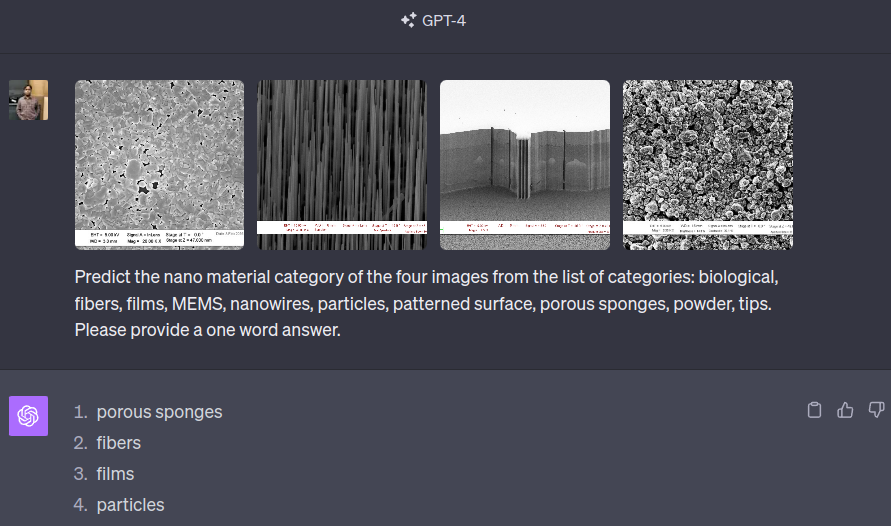}
\vspace{-2mm}
\caption{The electron micrographs shown above were provided as input to GPT-4V\cite{yang2023dawn} for nanomaterial categorization to determine how the multimodal model classifies nanomaterials in SEM images across different structural categories from a predefined list. However, the LMMs predictions were incorrect, with the actual nanomaterial categories being films, nanowires, MEMS and powder. It highlights the inherent limitations of visual processing capabilities of even advanced LMMs such as GPT-4V, reminding users to approach predictions with a degree of skepticism.}
\vspace{-4mm}
\label{fig:figure2}
\end{figure}

\vspace{0mm}
Recently, unimodal Large Language Models (LLMs) such as GPT-4 (language-only)\cite{gpt4} which are pre-trained autoregressive large-scale models on extensive, diverse text corpora in unsupervised learning settings following a fundamental paradigm ``prompt and predict" approach, have significantly transformed natural language processing(NLP), achieving improved performance across a wide range of NLP tasks, demonstrating better logical reasoning abilities, and generating human-like text. Zero-shot Chain of Thought(Zero-Shot CoT)\cite{wei2022chain} and Few-Shot (In-Context) learning(Few-Shot ICL)\cite{brown2020language} are prompt engineering strategies for designing and crafting tailored prompts for utilizing general-purpose LLMs in specialized language-based tasks or associated new, unseen problem-solving scenarios, thereby eliminating the need for traditional task-specific fine-tuning. Zero-Shot CoT relies on customized instructions without requiring explicit task-specific demonstrations(input-output pairs), requiring the language model to generalize from the implicit knowledge acquired during training to generate the output for the downstream task. Conversely, Few-shot ICL incorporates a few guiding demonstrations to learn from analogy along with the task-centric instructions to guide LLMs to generate the output simply by conditioning on the prompts. In recent times, OpenAI's GPT-4 with Vision (GPT-4V)\cite{gpt4v}, which possesses the ability to process and understand images, represents a significant advancement in the domain of large multimodal models (LMMs). It is more versatile than GPT-4, as it has broken the text-only barrier of previous language models, introducing visual understanding and analysis as a new dimension. GPT-4V is designed to accept multiple modalities, including both images and text as input, and generate text outputs. GPT-4V incorporates visual processing capabilities, enabling it to analyze image inputs provided by the user in conjunction with text, thereby facilitating visual question answering. Despite its advanced capabilities, when tested on SEM images\cite{aversa2018first} for nanomaterial categorization, GPT-4V incorrectly classified them, highlighting the limitations of LMMs. Figure \ref{fig:figure2} illustrates these limitations of GPT-4V.

\vspace{-4mm} % https://browse.arxiv.org/pdf/2306.14895.pdf

\begin{figure}[!ht]
\centering
\resizebox{1.125\linewidth}{!}{ 
\hspace*{0mm}\includegraphics[keepaspectratio,height=4.5cm,trim=0.0cm 0.0cm 0cm 0.1cm,clip]{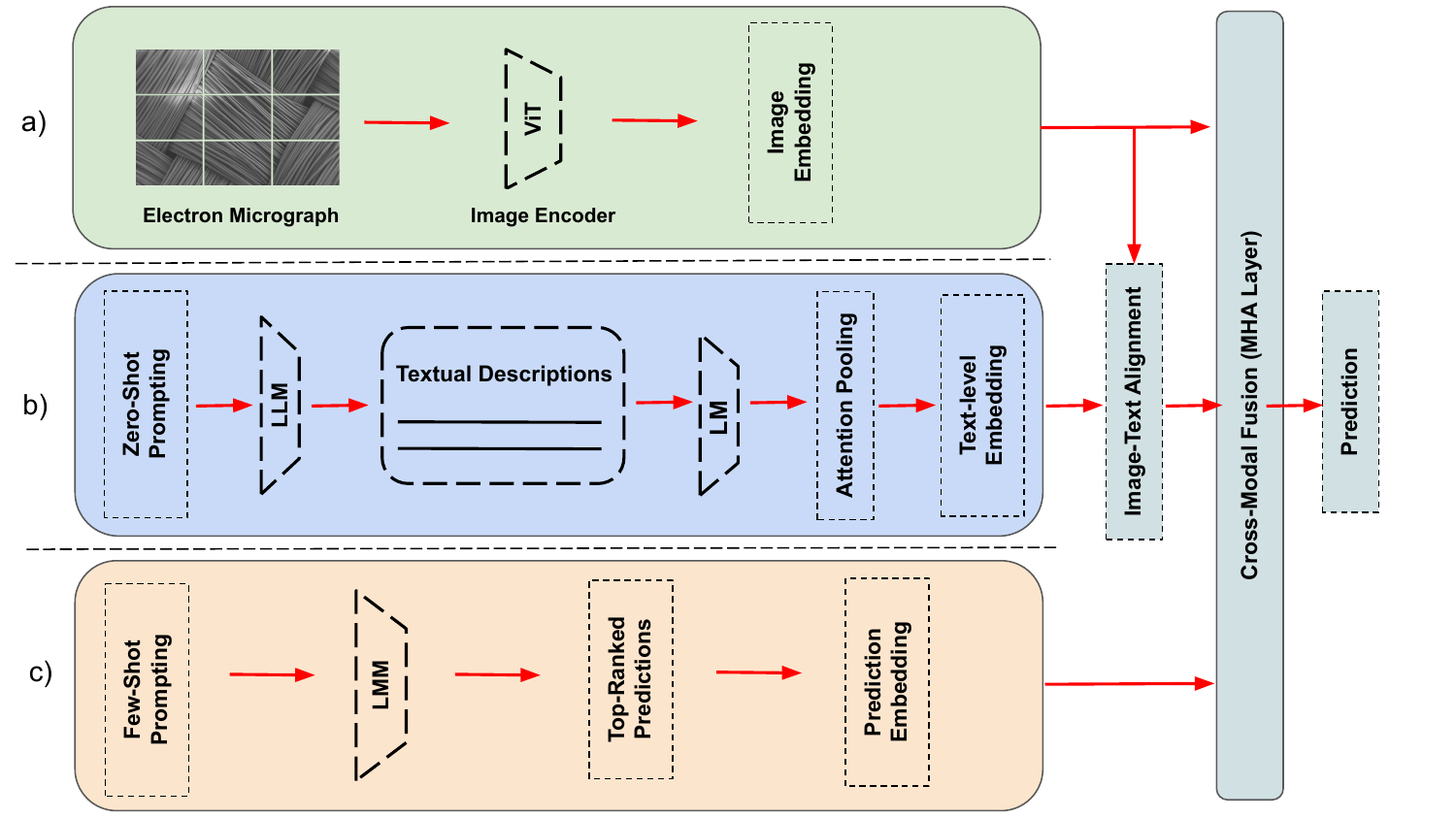} % left, bottom, right, top
}
\vspace{-6mm}
\caption{Our framework includes three methods: (a) Image Encoder (ViT), (b) Zero-Shot CoT prompting with LLMs, (c)  Few-shot prompting with LMMs, and (d) an output layer modeled with the multi-head attention (MHA) mechanism for integrating cross-domain embeddings and facilitating label prediction. }
\label{fig:figure3}
\vspace{-7mm}
\end{figure}

\vspace{-1mm}
Despite advances in the use of language-only LLMs such as GPT-4, LMMs like GPT-4V and other behemoths across various scientific disciplines, the synergistic integration of foundational LLMs and LMMs with computer vision algorithms in semiconductor research, particularly for the automated electron micrograph identification task, remains an underexplored approach in the evolution of intelligent semiconductor manufacturing processes. In this study, we introduce an automated nanomaterial identification framework, which is built upon cross-modal electron micrograph representation learning, referred to as \texttt{CM-EMRL} for brevity. Figure \ref{fig:figure3} illustrates the framework. The objective is to utilize the complementary strengths of LMMs, LLMs, and small-scale language models(LMs) to establish a more robust and accurate predictive framework. Closed-source LLMs like GPT-4\cite{gpt4}, while proficient in language understanding, have a black-box nature, and lack interpretability for downstream applications since they typically do not provide direct access to logits or token embeddings. In addition, their jack-of-all-trades approach makes them incredibly resource-intensive for repurposing and often poorly aligned with specialized tasks. On the other hand, open-source smaller LMs like BERT\cite{devlin2018bert} following ``pretraining and fine-tuning" approach, while cost-effective for repurposing through fine-tuning to align with specialized tasks and be interpretable, may fall short in terms of reasoning and generalization, often yielding less coherent and contextually relevant responses compared to LLMs. LMMs such as GPT-4V \cite{gpt4v} are more potent and versatile than language-only LLMs, as they incorporate multi-sensory capabilities that combine visual and language understanding. This enables users to instruct the multimodal model to analyze image inputs alongside textual information. Consequently, it offers the ability to perform complex tasks that require an understanding of both text and visual inputs, producing output that is contextually relevant to the combined data. LMMs excel in multimodal processing, with their remarkable capabilities to analyze images, identify objects, transcribe text, and decipher data, but they grapple with challenges related to interpretability, bias, unpredictability, and resource consumption. Navigating these challenges among LLMs, LMMs, and small-scale LMs demands a fine balance between performance, efficiency, and interpretability. Our study introduces a novel approach to the automatic nanomaterial identification task, combining the strengths of LMMs, LLMs, and small-scale LMs. The main contributions of this work can be summarized as follows:

\vspace{-1mm} 
\begin{itemize}
\item \textbf{Utilizing Vision Transformers (ViT) for Holistic Image Representation:} An input image is divided into patches treated as tokens, converted into 1D vectors, and enhanced with positional embeddings for location context. A classification token is added to achieve a global image representation. This token sequence is processed by a transformer architecture, specifically ViT\cite{dosovitskiy2020image}, to identify relationships between different image regions. The output corresponding to the classification token provides a comprehensive image representation. This appraoch employs the transformers to encapsulate the entire image visual context by treating the classification token's latent representation as an image-level embedding.
\vspace{-5mm} 
\item \textbf{Zero-shot Chain-of-Thought(CoT) LLMs Prompting and Cross-Modal Alignment:}Our study leverages powerful LLMs through Language Model as a Service (LMaaS), using their text-based inference APIs without accessing the parameters and gradients of the LLMs. Compared to fine-tuning task-specific LLMs, LMaaS efficiently deploys a single versatile LLM for various tasks using task-specific prompts. It optimizes tasks without backpropagation, ensuring low costs irrespective of the model size. The open-ended natural language CoT prompts guide LLMs to generate comprehensive textual descriptions of nanomaterials, covering their structure, synthesis methods, properties, and applications. After obtaining these technical descriptions, a smaller pretrained language model (LM) is used to summarize the LLM-generated content and compute high-level text embeddings by aligning them through supervised fine-tuning for the downstream nanomaterial identification task. Furthermore, in a cross-modal alignment scenario, a scaled dot-product attention mechanism matches image embeddings with their corresponding text-level embeddings. This mechanism calculates similarity scores to determine the best text match for a given image, ensuring robust alignment between different modalities. In brief, we utilize CoT LLM prompting to extract domain-specific knowledge and obtain image-aligned (nanomaterial-specific) text-level embeddings.
\vspace{-1mm} 
\item \textbf{In-Context Learning in LMMs via Few-Shot Prompting for Nanomaterial Identification:} Utilizing few-shot prompting can quickly adapt LMMs to perform new tasks, like nanomaterial identification, without extensive fine-tuning. By providing LMMs with a limited set of image-label pairs, these models can predict the category of unfamiliar or unseen nanomaterial images. Two strategies, random and similarity-driven sampling, influence the selection of these pairs. Instead of updating model parameters through supervised learning, this approach leverages the multimodal model's inherent knowledge, generating prediction embeddings by solely conditioned on the prompt.
\vspace{-5mm} 
\item \textbf{Unified Attention Layer:} We utilize a hierarchical multi-head attention mechanism to compute a cross-modal embedding from image-level, text-level, and prediction embeddings. This robust framework effectively integrates diverse information across these modalities, producing a holistic representation that can improve nanomaterial identification tasks. 
\vspace{-4mm}
\end{itemize}

\vspace{-3mm}
\section{Problem Statment}
\label{pm}
\vspace{-1mm}
Our study focuses on the classification of electron micrographs, an inductive learning challenge where the goal is to categorize previously unobserved micrographs by leveraging a labeled dataset, represented as $\mathcal{D}_L = (\mathcal{I}_L, \mathcal{Y}_L)$. We train a multi-modal encoder, defined by the non-linear transformation $g_{\gamma} : \mathcal{I} \rightarrow \mathcal{Y}$ on the labeled dataset in the context of supervised machine learning to predict the labels ($\mathcal{Y}_U$) for unlabeled micrographs ($\mathcal{I}_U$). Within this context, $\gamma$ denotes the trainable parameters, with the primary aim being to reduce the loss $\mathcal{L}_{\mathcal{I}}$, which is framed as:

\vspace{0mm}
\resizebox{0.90\linewidth}{!}{
\begin{minipage}{\linewidth}
\begin{equation}
 \min _{\gamma} \mathcal{L}_{\mathcal{I}}\left(\mathcal{I}_{i}, \gamma\right)=\sum_{\left(\mathcal{I}_{i}, y_{i}\right) \in \mathcal{D}_{L}} \ell\big(g_{\gamma}(\mathcal{I}_{i}), y_{i}\big)
 \end{equation}
 \end{minipage}
}

\vspace{-1mm}
where $y^{\text{pred}}_{i} = g_{\gamma}(\mathcal{I}_i)$ represents the predictions from the multi-modal encoder, $\ell(\cdot, \cdot)$ signifies the cross-entropy loss. 

\vspace{-2mm}

\vspace{-3mm}
\section{Proposed Method}
\label{pm}

\vspace{-1mm}
\paragraph{Electron Micrograph Encoder:} Let's consider an input image $\mathbf{I}$, which is represented as a 3D tensor with dimensions $H \times W \times C$, where $H$ represents the image's height in pixels, $W$ represents its width in pixels, and $C$ represents the number of channels of each pixel within the image. We divide the image into smaller, non-overlapping regions or patches to obtain a sequence of tokens. Each patch is treated as a token and has a fixed size with spatial dimensions of $P \times P \times C$, where $P$ denotes the patch size. The total number of patches is given by $n = \left(\frac{HW}{P^2}\right)$. We then linearly encode each patch, each of which has an overall size of $P^2C$, to flatten it into a 1D vector represented as $\mathbf{I'} \in \mathbb{R}^{n \times d}$, where $d$ is the patch embedding dimension. To provide the model with spatial information, we add positional embeddings to each patch. These positional embeddings are learnable vectors representing the position of each patch within the image grid. They help us understand the relative positions of different patches, and we add the position embeddings element-wise to the patch embeddings. In addition, we append a classification token $<\hspace{-1mm}\textit{cls}\hspace{-1mm}>$ to the patch sequence. This token aggregates information from all patches and provides an embedding of the entire patch sequence, creating a global representation of the entire input image. We input this augmented token sequence into ViT\cite{dosovitskiy2020image}, which consists of multiple stacked transformer encoder layers. Each encoder layer processes the patch embeddings hierarchically, refining representations at different abstraction levels. After passing through the transformer layers, we consider only the output embedding $h_{\textit{cls}}$ corresponding to the $<\hspace{-1mm}\textit{cls}\hspace{-1mm}>$ token as the representation of the entire image. This embedding aggregates information from all the patches and summarizes it. In summary: (a) We split the input image into patches, flatten them into 1D vectors, augment them with positional embeddings to provide spatial information, and include a classification token $<\hspace{-1mm}\textit{cls}\hspace{-1mm}>$. (b) We process the resulting sequence of patch embeddings using a transformer-based architecture to capture long-range dependencies and relationships between different regions of the electron micrograph.

\vspace{-1mm}
\paragraph{Zero-Shot CoT LLMs Prompting:} We access LLMs through the LMaaS\cite{sun2022black} hosted by the cloud-based service provider,  which provides a text-based black-box API interaction to send text inputs and receive generated text without access to the underlying model parameters or gradients or the model's internal mechanisms. We utilize open-ended natural language prompts, designed to be flexible and non-restrictive, to instruct LLMs to generate the detailed technical descriptions related to various aspects of nanomaterials, including their structure, properties, and applications. We query LMMs to generate detailed descriptions of nanomaterials by employing a customized prompt template for zero-shot generation tasks, steering them through CoT prompts in a zero-shot context. This process involves extracting pre-existing domain-specific knowledge embedded within the language model parameters acquired during training to generate in-depth, technical descriptions of nanomaterials that are both detailed and comprehensive. The customized CoT prompt format is as follows:

\vspace{-2.5mm}
\begin{tcolorbox}[colback=white!5!white,colframe=black!75!black]% [width=10cm]
\vspace{-2mm}
\textbf{Prompt 1:} Introduction: Provide an overview of the nanomaterial category and its significance across various fields.
\textbf{Prompt 2:} Definition and Structure: Define the nanomaterial category and describe its typical structure at the nanoscale.
\textbf{Prompt 3:} Synthesis Methods: Examine different methods employed for synthesizing or fabricating nanomaterials within this category. Discuss both their advantages and limitations.
\textbf{Prompt 4:} Properties: Highlight the unique physical, chemical, and electronic properties exhibited by nanomaterials in this category. Explain how these properties differ from those of bulk materials.
\textbf{Prompt 5:} Surface Modification: Describe strategies used to modify the surface properties of nanomaterials in this category, including techniques like functionalization, coating, or doping. Explain how these modifications enhance their performance or enable specific applications.
\textbf{Prompt 6:} Applications: Explore the extensive range of applications wherein nanomaterials from this category find use. Discuss their potential impact on fields such as electronics, energy, medicine, and more.
\vspace{-2mm}
\end{tcolorbox}

\vspace{-1.5mm}
The structured prompts are designed to facilitate a comprehensive, in-depth exploration of various facets, ranging from fundamental properties to practical applications and potential risks associated with these nanomaterials. Prompting LLMs generates text that responds to and elaborates on the specific aspects mentioned in each prompt.

\vspace{-1.5mm}
\begin{tcolorbox}[colback=white!5!white,colframe=black!75!black]%[width=10cm]
\vspace{-1.5mm}
\centering
(\textbf{Chatbot's Response}) [Generated Text]
\vspace{-1.5mm}
\end{tcolorbox}

\vspace{-1.5mm}
Table \ref{tab:lmprompts} shows the sample responses generated by GPT-4 to the natural language prompts on MEMS devices. In the subsequent section, we will describe our approach to integrate technical descriptions into a small-scale LM for fine-tuning on the supervised nanomaterial identification task.

\vspace{-2.5mm} 
\paragraph{Fine-Tuning Smaller LMs:} Our approach employs a smaller pretrained language model (LM) to summarize the technical descriptions generated by a large language model (LLM) on nanomaterials. In our study, we utilize a pre-trained small-scale LM, DeBERTa\footnote{For more information, refer to the DeBERTa model documentation available at \url{https://huggingface.co/docs/transformers/index}.}\cite{he2020deberta}, an improved version of the BERT architecture. We fine-tune this small-scale LM on the generated technical descriptions for domain-specific customization on the downstream task. This helps the language model learn the statistical relationships between words and phrases in the  large corpus of LLM textual outputs, thereby facilitating the generation of context-aware token embeddings. We input the text sequences generated by LLMs (denoted as $\mathcal{S}_\textrm{expl}$) into the $\textrm{LM}_{\textrm{expl}}$ model, which then generates expressive token embeddings as follows:

\vspace{0mm}
\resizebox{0.965\linewidth}{!}{
\hspace{0mm}\begin{minipage}{\linewidth}
\begin{equation}
h_{\textrm{expl}} = \textrm{LM}_\textrm{expl}(s_{\textrm{expl}})
\end{equation}
\end{minipage}
} 

\vspace{1mm}
where the deep contextualized embeddings are denoted as $h_{\textrm{expl}} \in \mathbb{R}^{\hspace{0.5mm}m \times d}$, where $m$ represents the number of tokens in $\mathcal{S}_{\textrm{expl}}$ and $d$ is token embedding dimension.  We perform weighted average of all token embeddings for a comprehensive representation of the entire text. We use a softmax  attention mechanism to compute interpretable attention weights, denoted as $\alpha$, and subsequently use these weights to sum-pool the token embeddings, encoding the textual descriptions into a fixed-size, high-level text representation as follows:

\vspace{0mm}
\resizebox{0.95\linewidth}{!}{
\hspace{0mm}\begin{minipage}{\linewidth}
\begin{equation}
\alpha = \mbox{softmax}(q); \hspace{2mm} q = \mathbf{u}^Th_{\textrm{expl}}
\end{equation}
\end{minipage}
} 

\vspace{0mm}
\resizebox{0.915\linewidth}{!}{
\hspace{0mm}\begin{minipage}{\linewidth}
\begin{equation}
h_{\text{text}} = \sum_{j=0}^m{\alpha_i h^{(j)}_{\textrm{expl}}}
\end{equation}
\end{minipage}
} 

\vspace{0mm}
where $\mathbf{u}$ is a learnable vector and the subscript $j$ refers to token. The text-level embedding $h_{\text{text}} \in \mathbb{R}^{(d)}$ encapsulates the relevant and concise information at the core of the domain knowledge in technical descriptions, which is extracted from the general-purpose LLMs for each nanomaterial. 

\vspace{-2mm}
\paragraph{Cross-Modal Alignment Using Multi-Head Self-Attention:}
We employ the multi-head self-attention mechanism to align image embeddings with their corresponding text embeddings for image-text matching purposes. This approach emphasizes specific aspects or features of the image that is semantically relevant to the textual descriptions, taking into consideration the different facets of the relationship of the cross-domain modalities (both text-level and image embeddings). The mechanism calculates similarity scores between the provided image embedding and all text embeddings. The text embedding with the highest similarity score is considered the best match for the given image embedding. We initially concatenate the text-level embeddings for the different nanomaterial categories to obtain a unified text-level embedding $\mathbf{h}_\text{text} = [ \mathbf{h}^{(1)}_\text{text}, \cdots, \mathbf{h}^{(c)}_\text{text}]$, where c refers to the total number of nanomaterials. We compute the value and key projections for the unified text-level embedding, which represents the combined semantic information of all nanomaterials, for each head $\text{h}$ as follows:

\vspace{-2mm}
\resizebox{0.95\linewidth}{!}{
\begin{minipage}{\linewidth}
\begin{align}
K^h_{\text{text}} &= \mathbf{h}_\text{text} \text{W}^h_{K_\text{text}}; V^h_{\text{text}} = \mathbf{h}_\text{text} \text{W}^h_{V_\text{text}} \nonumber
\end{align}
\end{minipage}
}

Similarly, the query projection for image embedding $\mathbf{h}_{\textit{cls}}$ for each head $\text{h}$ is as follows:

\vspace{-2mm}
\resizebox{0.95\linewidth}{!}{
\begin{minipage}{\linewidth}
\begin{align}
Q^h_{\textit{cls}} &= \mathbf{h}_{\textit{cls}} \text{W}^h_{Q_{\textit{cls}}} \nonumber
\end{align}
\end{minipage}
}

where $\text{W}^h_{K_\text{text}}$, $\text{W}^h_{V_\text{text}}$, and $\text{W}^h_{Q_{\textit{cls}}}$ are trainable weight matrices. We now utilize the scaled-dot product attention mechanism\cite{vaswani2017attention} to compute the normalized attention score, which measures the semantic similarity between the image embedding and each text embedding for a specific attention head $\text{h}$ as follows:

\vspace{-2mm}
\resizebox{0.95\linewidth}{!}{
\begin{minipage}{\linewidth}
\begin{align}
\text{A}_{w}^h = \text{softmax}\left( \frac{Q^h_{\textit{cls}} (K^h_{\text{text}})^\top}{\sqrt{d_k}} \right) \nonumber
\end{align}
\end{minipage}
} 

where \( d_{k} \) denotes the dimensionality of the keys. The attention weights align complementary information from the cross-domain embeddings, focusing on relevant information for cross-modal alignment and capturing richer semantics by allowing the framework to dynamically weigh different parts of the input based on their relevance to the context. We then compute the weighted sum of the value projection as follows:

\vspace{-3mm}
\resizebox{0.95\linewidth}{!}{
\begin{minipage}{\linewidth}
\begin{align}
\text{O}^h_{\text{text}} = \text{A}_{w}^h V^h_{\text{text}} \nonumber
\end{align}
\end{minipage}
}

\vspace{1mm}
We concatenate the outputs across different heads because it encapsulates perspectives from multiple heads which focus on different aspects or features, thus producing a more comprehensive and robust alignment between the two modalities. We project the outputs to obtain the final representation as follows:

\vspace{-5mm}
\resizebox{0.95\linewidth}{!}{
\begin{minipage}{\linewidth}
\begin{align}
\text{O}_{\text{text}} = \left[ \text{O}^{1}_{\text{text}}, \ldots, \text{O}^{H}_{\text{text}} \right] \text{W}_O \nonumber
\end{align}
\end{minipage}
}

\vspace{1mm}
where \( \text{H} \) represents the total number of heads and $\text{W}_O$ deontes the trainable weight matrix. We now compute the cosine similarity between the two cross-domain embeddings as follows:

\vspace{-5mm}
\resizebox{0.95\linewidth}{!}{
\begin{minipage}{\linewidth}
\begin{align}
\text{Sim} = \frac{\text{O}_{\text{text}} \cdot \mathbf{h}_{\textit{cls}}}{||\text{O}_{\text{text}}||_2 \times ||\mathbf{h}_{\textit{cls}}||_2}  \nonumber
\end{align}
\end{minipage}
}

\vspace{1mm}
where $\text{sim} \in \mathbb{R}^{c}$, and we select the best match based on the highest similarity value. The index is determined for the text-level embedding with the highest similarity score as follows:

\vspace{-6mm}
\resizebox{0.95\linewidth}{!}{
\begin{minipage}{\linewidth}
\begin{align}
i^* = \text{argmax}_i \left( \text{Sim} \right) \nonumber
\end{align}
\end{minipage}
}

\vspace{1mm}
Here, \( i^* \)  is the index of the best-matching text-level embedding. So, the best-matching text-level embedding would be:

\vspace{-5mm}
\resizebox{0.95\linewidth}{!}{
\begin{minipage}{\linewidth}
\begin{align}
\mathbf{h}^*_{\text{text}} = \mathbf{h}^{(i^*)}_{\text{text}} \nonumber
\end{align}
\end{minipage}
}

\vspace{1mm}
This is essentially a matching mechanism that seeks to find the best pairwise alignment among the various text-level embeddings and the image embedding. We utilize backpropagation error in the downstream supervised multi-classification task to fine-tune the ViT and smaller LMs to maximize the pairwise alignment between the complementary image embedding ($\mathbf{h}_{\textit{cls}}$) and its corresponding text-level embedding ($\mathbf{h}^*_{\text{text}}$). To summarize, $\mathbf{h}^*_{\text{text}}$ incorporates the prior knowledge obtained from LLMs for the appropriate nanomaterial underlying the elecron micrographs as auxiliary information to support multi-modal learning.

\vspace{-1mm}
\paragraph{Few-Shot LMM Prompting:} Few-shot prompting enables LMMs such as GPT-4V to adapt to new tasks without the need for explicit, gradient-based fine-tuning\cite{brown2020language} using the labeled data. This approach allows LMMs to learn by analogy, utilizing only a few input-output pairs specific to the downstream task. Few-shot prompting leverages the implicit knowledge embedded in pretrained LMM parameters to adapt to new tasks through task-specific demonstrations, thereby avoiding the need to repurpose LMMs with parameter updates. The context-augmented prompt provides task-specific instructions and demonstrations(input-output pairs), enabling LMMs to generate outputs conditioned on the prompt for improved generalization performance on the new, unfamiliar tasks. In the case of nanomaterial identification tasks, few-shot prompting involves creating a context-augmented prompt using a few input-output mappings $(\mathcal{I}_i, \mathcal{Y}_i)$, which are a small number of image-label pairs as demonstrations sampled from the training data relevant to the query(new/unseen) image. These mappings provide relevant context to aid in understanding and classifying unseen images. The task-specific instruction is related to the query image, instructing the multimodal model to predict its associated label.  At inference time, given test images \( \mathcal{I}_{\text{test}} \), few-shot prompting predicts an output label based on the conditional probability distribution, \( \mathbf{P}(\mathcal{Y}_{\text{test}} \mid ((\mathcal{I}_{\text{train}}, \mathcal{Y}_{\text{train}}), \mathcal{I}_{\text{test}})) \). To explore how the quality and quantity of few-shot demonstrations affect the performance in nanomaterial identification tasks, we consider two distinct sampling strategies: ``Random" and ``Similarity-driven Sampling". The random approach involves selecting demonstrations (image-label pairs) arbitrarily from the training data without any specific criteria, serving as a baseline for evaluation. In contrast, the similarity-driven sampling strategy employs cosine similarity to identify the top-$K$ images that most closely resemble a given query image within the training data. This strategy operates under the hypothesis that demonstrations which are more representative of the query image's data distribution can potentially enhance model adaptability and accuracy. By utilizing both the diverse strategies to construct augmented prompts, we aim to provide a comprehensive analysis of how different demonstration sampling methods impact the effectiveness of few-shot learning in nanomaterial identification tasks. Furthermore, the efficacy of the demonstrations is inherently related to the sampling methods used to retrieve the top-\( K \) images that align most closely with the query image. To further explore the impact of the volume of demonstrations on performance, we adjust the number of demonstrations \( K \) for each query image. In summary, our objective is to provide LMMs with a context-augmented prompt, comprising image-label pairs selected from the training data, along with task-specific instructions that guide the LMMs in predicting the nanomaterial category of the query image. This task evaluates the LMMs' ability to predict nanomaterial categories based on the prompt without any parameter updates, distinguishing it from traditional supervised learning, where models are fine-tuned on labeled data. For each query image, the LMMs generate a \( c \)-dimensional one-hot vector \( h_{\text{pred}} \in \mathbb{R}^{\hspace{0.25mm}c} \), where \( c \) denotes the predefined number of nanomaterial categories. This vector is linearly encoded into a high-dimensional space to produce a prediction embedding \( h_{\text{ICL}} \in \mathbb{R}^{\hspace{0.25mm}d} \), which encapsulates the LMMs predictions. Here, \( d \) represents the embedding dimension and \( c \ll d \). An example of an ICL prompt is as follows,

\vspace{-2mm}
\begin{tcolorbox}[colback=white!5!white,colframe=black!75!black]% [width=10cm]
\centering
\vspace{-2mm}
Below are the input-output pairs (image-label pairs) for the nanomaterial identification task. Predict the nanomaterial category for the query image.
\vspace{-2mm}
\end{tcolorbox}

\vspace{-1mm}
The prediction embedding likely contains valuable information about potential outcomes, allowing the framework to refine its cross-modal representation for better alignment with desired results. Given the complexity of nanomaterials structures and properties, this prediction embedding has the potential to capture some of that complexity, guiding the framework toward correct identification through the integration of prior knowledge and auxiliary information. A general purpose GPT-4V is accessible to ChatGPT Plus subscribers with a usage cap at chat.openai.com. However, it's not currently available for public use through Multimodal Modeling as a Service (MMaaS) — a cloud-based service that accepts both image and textual inputs as prompts. By utilizing black-box GPT-4V through MMaaS as an on-demand service, typically hosted on cloud servers and accessed via an API, users can design task-specific prompts to query pre-trained LMMs for solving multimodal tasks of interest. This is analogous to how users might access LLMs via Language Modeling as a Service (LMaaS\cite{sun2022black}) to tackle language-specific tasks. GPT-4V is anticipated to become publicly accessible by mid-November 2023. APIs are designed for large-scale and concurrent requests and are ideal for integration into automated systems. Conversely, websites might not efficiently handle numerous interactions in rapid succession, and automating tasks on them could be prohibited. Manually sending inputs for GPT-4V for the entire training dataset would be a daunting and tedious task. Instead, we select a subset of images from the whole training dataset, termed `Confounding or Ambiguous Micrographs', for few-shot prompting of GPT-4V. The selection process for these images is as follows: The SEM electron micrographs, initially sized at $1024 \times 768 \times 3$ pixels, were downscaled to $224 \times 224 \times 3$ pixels. They were then normalized using z-score normalization and flattened into one-dimensional vectors. Their dimensionality was further reduced using Principal Component Analysis (PCA) before employing the K-Means clustering algorithm. We chose $K=10$ clusters based on a predefined number of nanomaterial categories based on benchmark dataset. This method identifies images that are challenging to classify by measuring distances from centroids, assessing cluster variance, and calculating the Silhouette Score. Ground truth labels aid in the evaluation of the clustering's effectiveness. We sampled a fixed \(10\%\) of the ambiguous images from the entire dataset. For these images, we apply few-shot prompting of GPT-4V to predict labels. The goal is to learn the optimal projection layer weight matrices, which transform the GPT-4V predictions (one-hot vectors) into a high-dimensional space, producing a prediction embedding \( h_{\text{ICL}} \in \mathbb{R}^{d} \) that encapsulates the GPT-4V predictions. The projection layer is subsequently trained through the supervised learning task. This training aims to minimize the cross-entropy loss and obtain optimal weights.

\vspace{-2mm}
\paragraph{Unified Attention Layer:} We compute the cross-modal embedding, denoted as $h_{\text{cross}}$, using a hierarchical multi-head attention mechanism that integrates the original image embedding $\mathbf{h}_{\text{cls}}$, the text-level embedding $\mathbf{h}^*_{\text{text}}$, and the prediction embedding $\mathbf{h}_{\text{ICL}}$. This mechanism offers a robust framework for integrating diverse information from different modalities. As a result, it can produce a more holistic representation that encompasses a wide range of information, potentially improving the performance of nanomaterial identification tasks. In general, the multi-head attention mechanism uses multiple heads to capture different attention patterns, allowing the model to recognize a variety of relationships in the data from multiple subspace representations. Given queries Q, keys K, and values V, the scaled dot-product attention is defined as follows:

\vspace{-6mm}
\resizebox{0.95\linewidth}{!}{
\begin{minipage}{\linewidth}
\begin{align}
\text{Attention}(\mathbf{Q}, \mathbf{K}, \mathbf{V}) = \text{softmax}\left(\frac{\mathbf{Q}\mathbf{K}^T}{\sqrt{d_k}}\right) \mathbf{V} \nonumber
\end{align}
\end{minipage}
}

where \( d_k \) is the dimensionality of the keys. The \textit{multi-head} attention mechanism employs multiple heads to integrate various attention patterns into a unified representation. Each of these heads utilizes the scaled dot-product attention on distinct linear transformations of the input queries, keys, and values. These transformations are parameterized by matrices \( \mathbf{W}_{Q_i} \), \( \mathbf{W}_{K_i} \), and \( \mathbf{W}_{V_i} \). The final output is derived by concatenating the results from these heads and subjecting it to a subsequent matrix transformation \( \mathbf{W}_O \).

\vspace{-2mm}
\resizebox{0.95\linewidth}{!}{
\begin{minipage}{\linewidth}
\begin{align}
\text{MultiHead}(\mathbf{Q}, \mathbf{K}, \mathbf{V}) &= \text{Concat}(\text{head}_1, ..., \text{head}_h) \mathbf{W}_O \nonumber \\
\text{where} \quad \text{head}_i &= \text{Attention}(\mathbf{Q}\mathbf{W}_{Q_i}, \mathbf{K}\mathbf{W}_{K_i}, \mathbf{V}\mathbf{W}_{V_i}) \nonumber
\end{align}
\end{minipage}
}

\vspace{1mm}
Given the context described earlier, the unified attention layer employs the multi-head attention mechanism in a hierarchical fashion to derive the cross-modal embedding. The procedure consists of two main stages: (1) \textbf{Image-Text Attention:} Here, the unified attention layer focuses on the image embedding \( \mathbf{h}_{\text{cls}} \) in relation to the text-level embedding \( \mathbf{h}^*_{\text{text}} \). The result is an intermediate embedding, \( \mathbf{h}_{\text{img-text}} \), which amalgamates details from both image and text modalities through the multi-head attention mechanism. The primary intent of this step is to incorporate relevant textual information guided by the image's context. This can be mathematically described as follows:

\vspace{-4mm}
\begin{equation}
\mathbf{h}_{\text{img-text}} = \text{MultiHead}(\mathbf{h}_{\text{cls}}, \mathbf{h}^*_{\text{text}}, \mathbf{h}^*_{\text{text}})  \nonumber
\end{equation}

\vspace{-1mm}
(2) \textbf{Image-Text-Prediction Attention:} During this stage, the previously derived intermediate embedding \( \mathbf{h}_{\text{img-text}} \) undergoes further refinement. The unified attention layer aligns this embedding with the prediction embedding \( \mathbf{h}_{\text{ICL}} \), computing the final cross-modal representation \( h_{\text{cross}} \). This stage aims to combine insights from the intermediate representation with the prediction embedding, creating a comprehensive representation that seamlessly integrates various modalities. This can be represented mathematically as:

\vspace{-3mm}
\begin{equation}
h_{\text{cross}} = \text{MultiHead}(\mathbf{h}_{\text{img-text}}, \mathbf{h}_{\text{ICL}}, \mathbf{h}_{\text{ICL}}) \nonumber
\end{equation}

\vspace{-1mm}
In summary, the unified attention layer uses multi-head attention mechanisms to capture and integrate information from multiple different modalities (image, text, prediction) in a hierarchical manner, resulting in the comprehensive cross-modal embedding that can be used for nanomaterial identification tasks. This mechanism employs multiple sets of learned weight matrices to emphasize various aspects or relationships within the data. Consequently, this approach has the potential to foster robust and enriched embeddings capable of capturing complex patterns. Additionally, it aids in focusing on contextually relevant information and in achieving semantic alignment across different embeddings, thereby enhancing the capacity to identify and utilize important features in the input data. Finally, we linearly transform the final unified cross-modal embedding to obtain a probability distribution $\text{p}_{i}$ over the possible outcomes, as follows:

\vspace{-3mm}
\resizebox{0.95\linewidth}{!}{
\hspace{0mm}\begin{minipage}{\linewidth}
\begin{align}
\text{p}_{i} &= \text{softmax}\big(\text{W} h_{\text{cross}} \big)   \nonumber
\end{align}
\end{minipage}
} 

\vspace{0mm}
$\text{p}_{i}$ represents the probability distribution across nanomaterial categories. We apply the argmax operation to $\text{p}_{i}$ to determine the most likely nanomaterial category predicted by the framework. In our approach, we concurrently carry out three tasks: (i) we compute image embeddings using an image encoder, (ii) we utilize zero-shot CoT prompting with LLMs to generate technical descriptions of nanomaterials and fine-tune smaller pre-trained LMs on the generated descriptions—subsequently, a softmax attention pooling mechanism is employed to produce text-level embeddings, and (iii) we employ few-shot prompting of LMMs to derive prediction embeddings. Following this, we jointly optimize these different embeddings using a hierarchical multi-head attention mechanism for supervised learning tasks. The overarching goal is to reduce the cross-entropy loss and improve multi-class classification accuracy. Furthermore, the MHA adeptly captures and aligns diverse data sources, making it indispensable for multi-modal integration and analysis. This capability is especially crucial in fields like nanomaterial analysis, where multiple modalities offer complementary insights.

\vspace{-3mm} 
\section{Experiments And Results}

\begin{figure}[htbp]
\centering
     \subfloat{\hspace{-0mm}\includegraphics[width=0.11\textwidth]{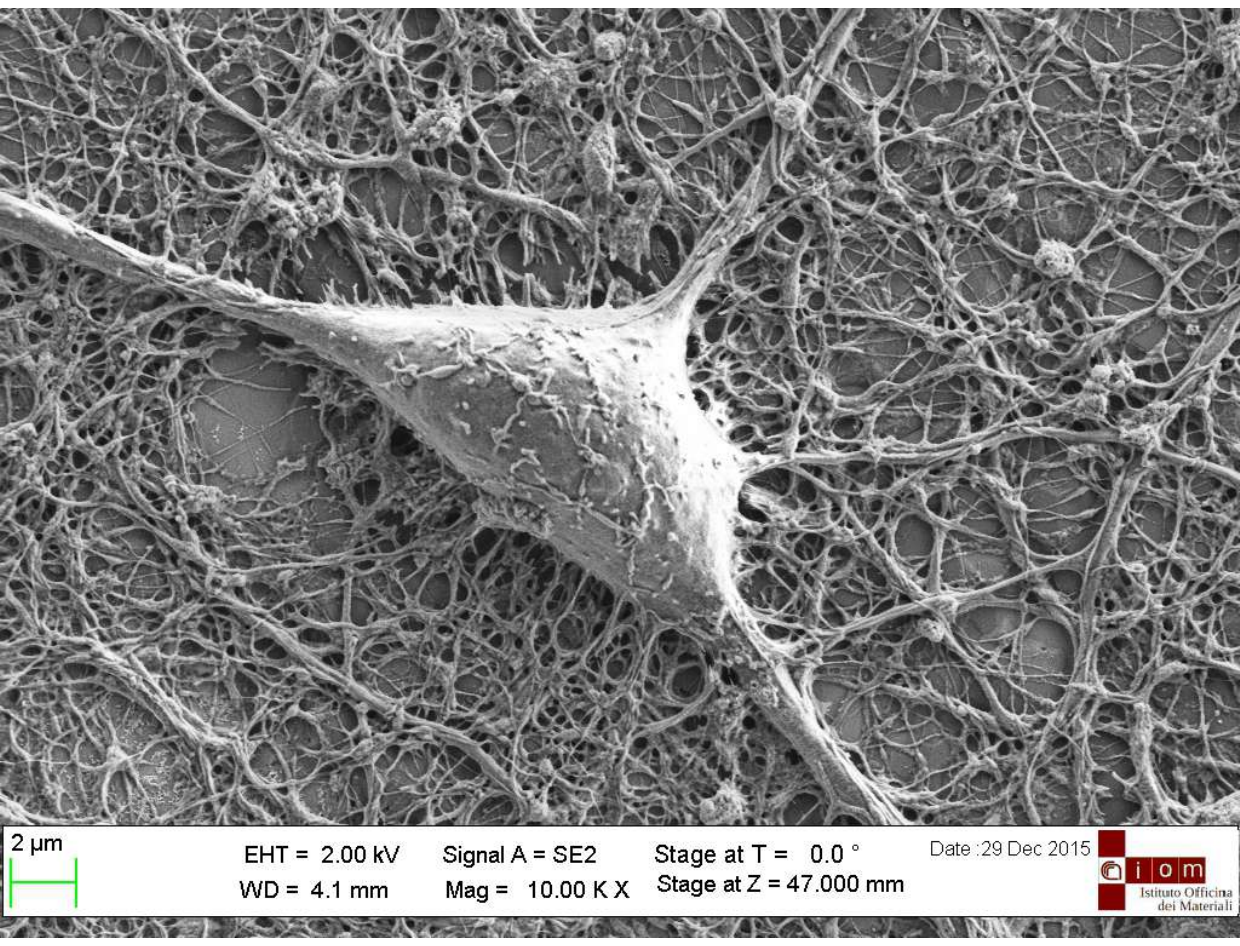}
     \includegraphics[width=0.11\textwidth]{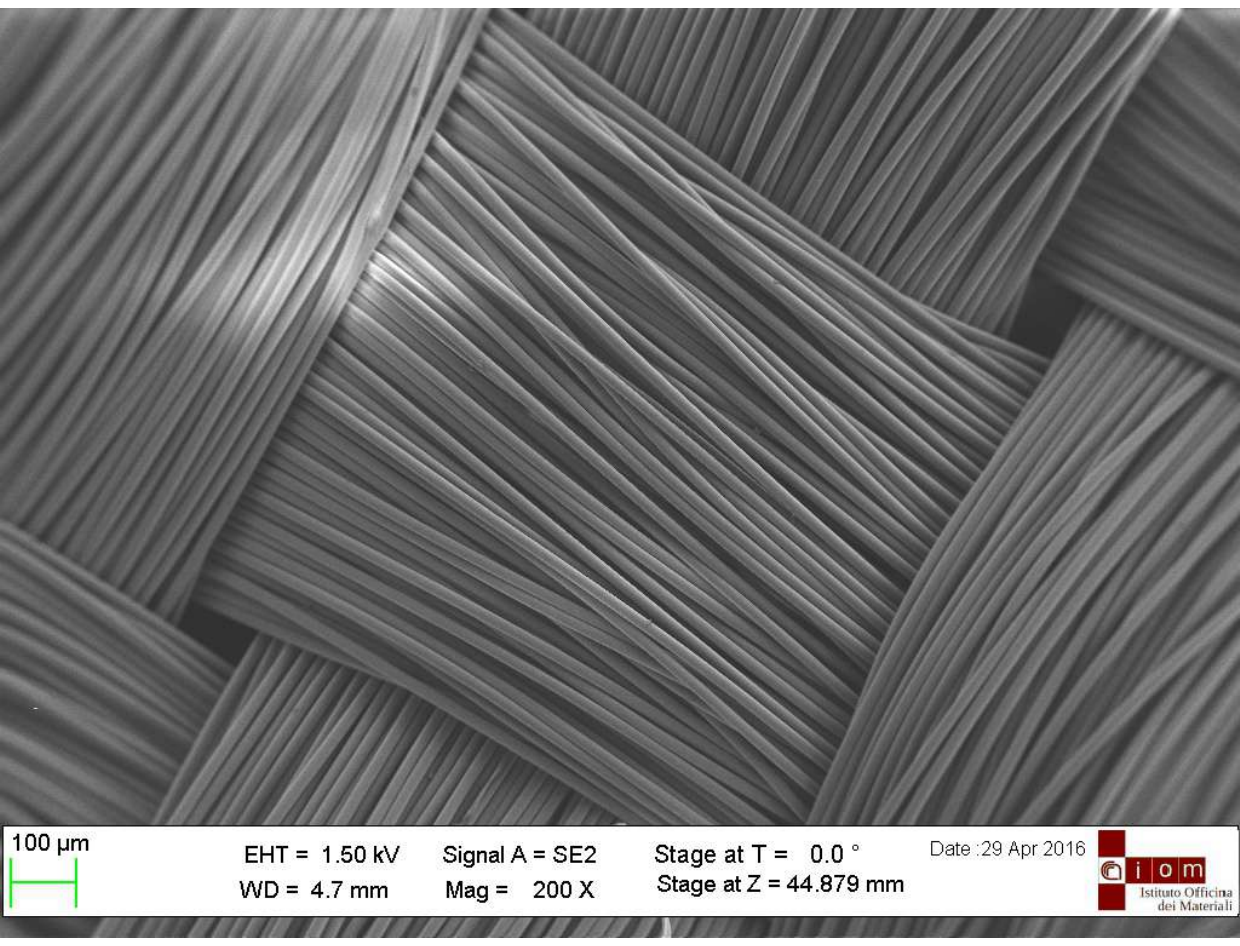}
     \includegraphics[width=0.11\textwidth]{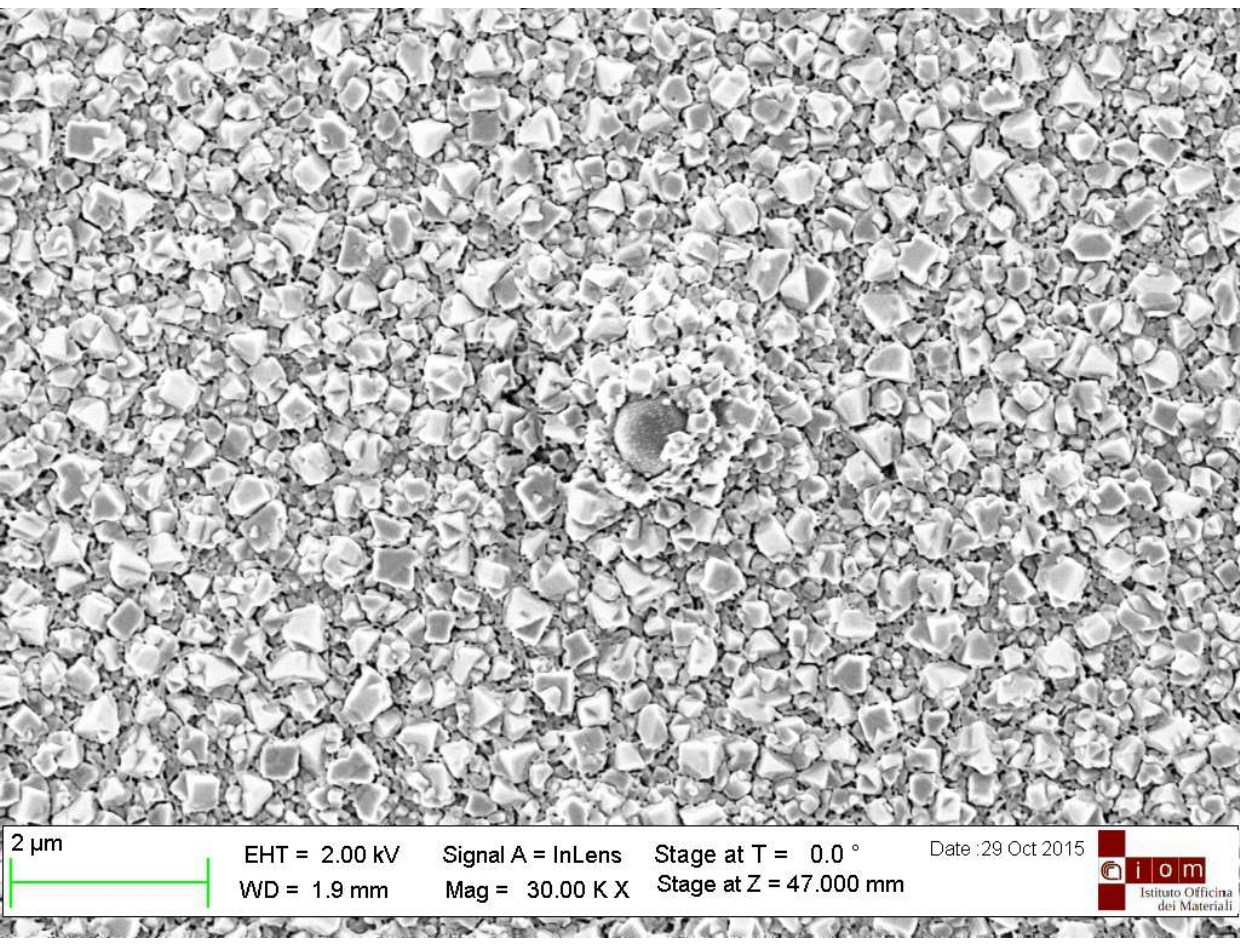}
     }
          \vspace{-12.5mm}
     \qquad
     \subfloat{\hspace{-0mm}\includegraphics[width=0.11\textwidth]{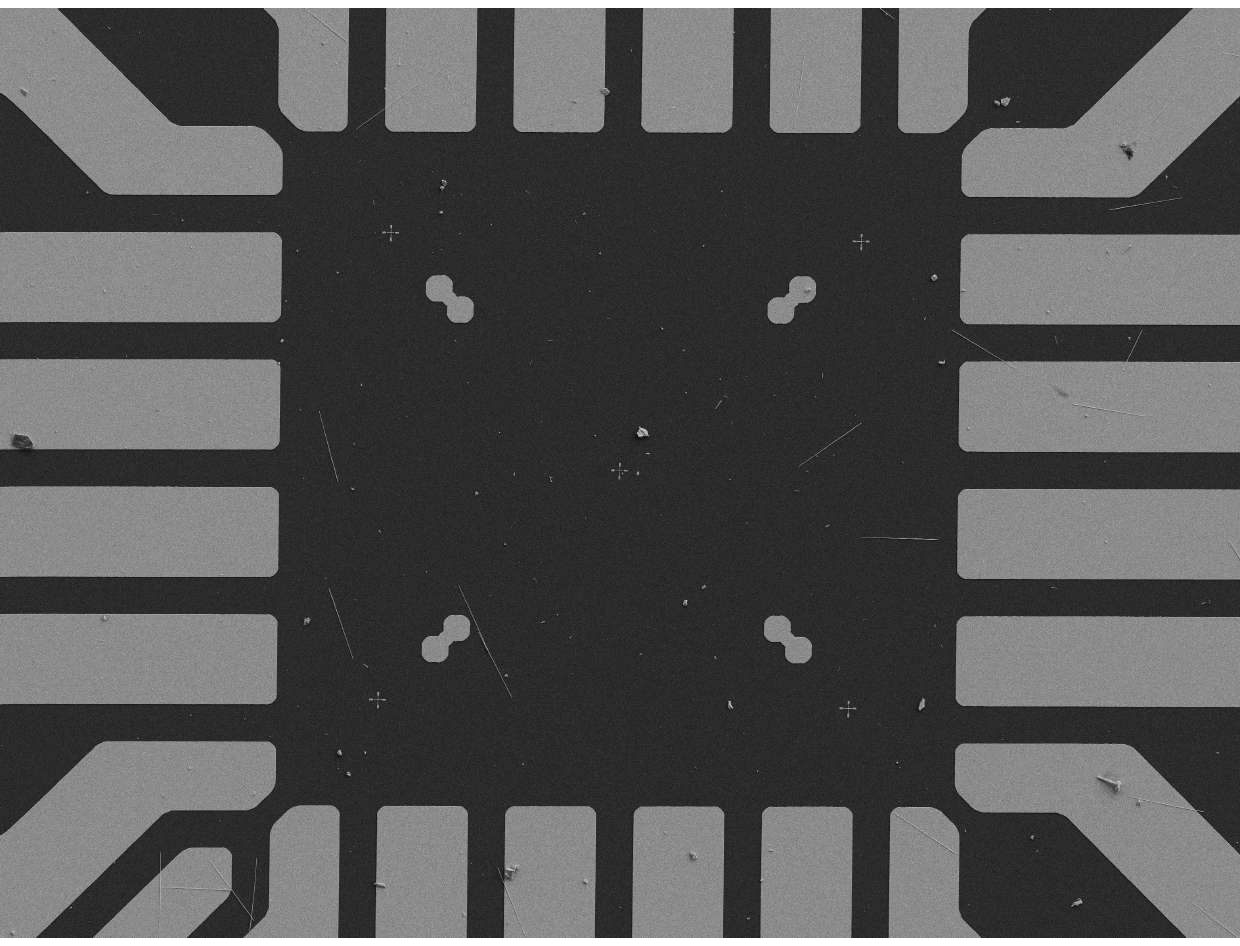}
     \includegraphics[width=0.11\textwidth]{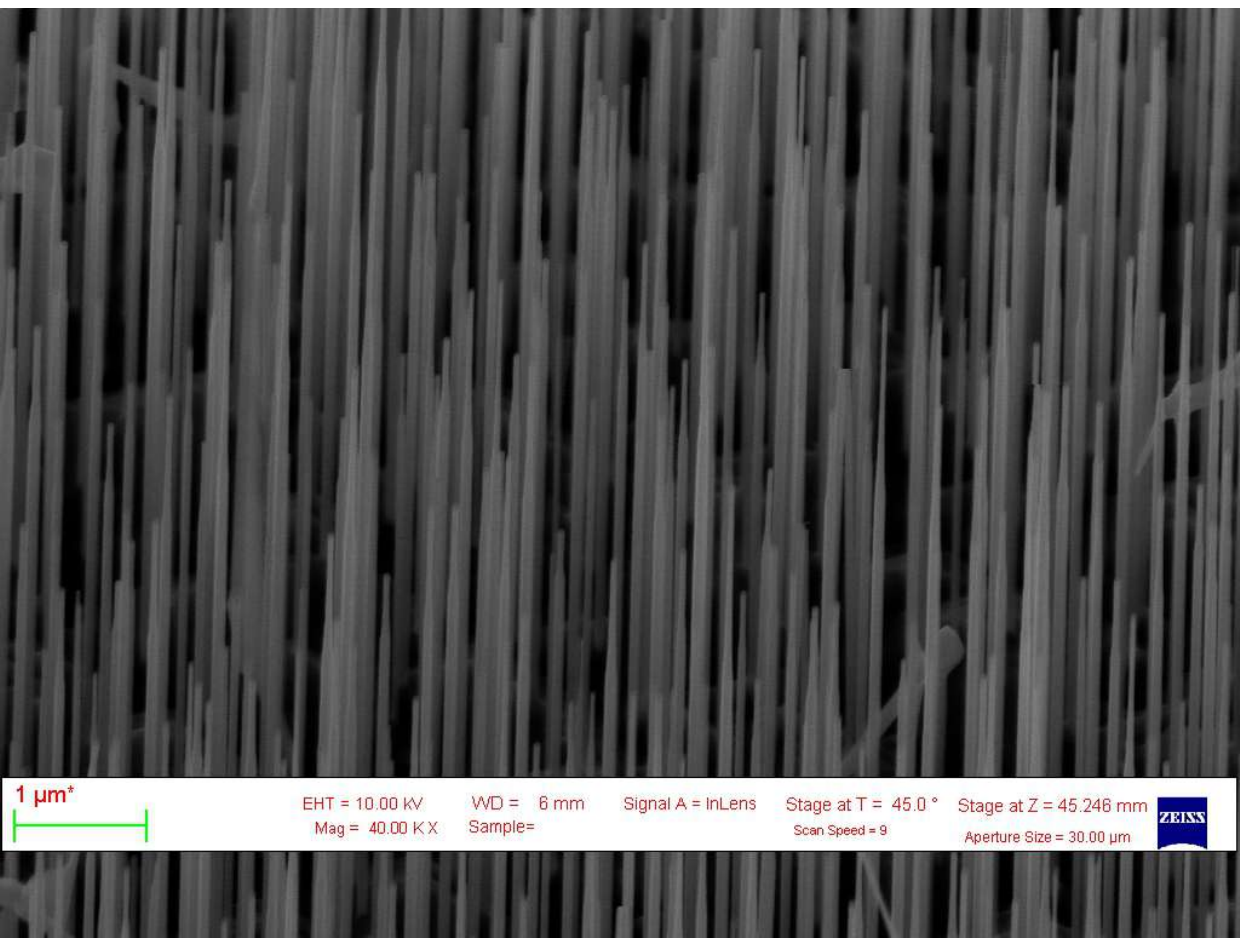}
     \includegraphics[width=0.11\textwidth]{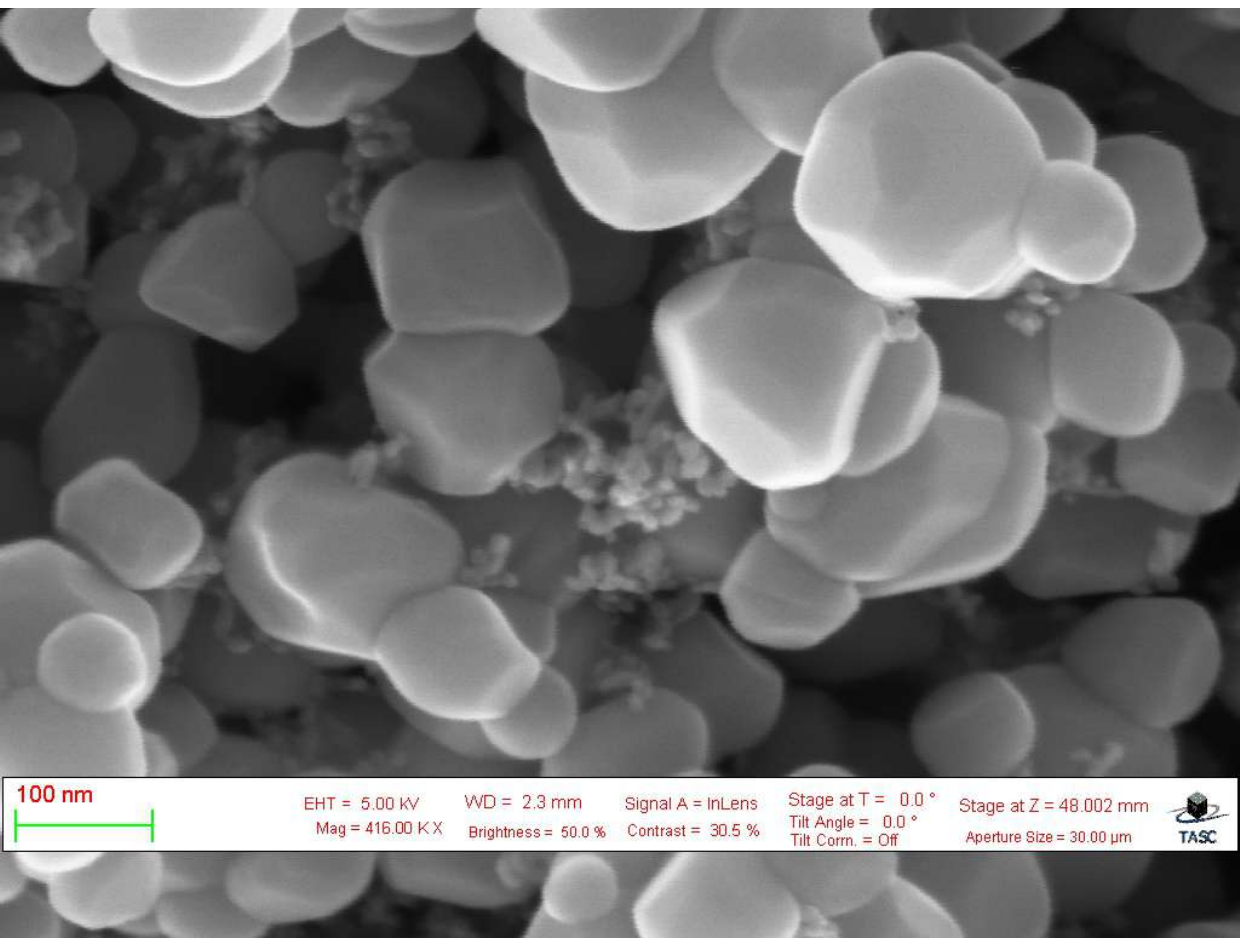}
     }
     \vspace{-13mm}
     \qquad
     \subfloat{\hspace{-0mm}\includegraphics[width=0.11\textwidth]{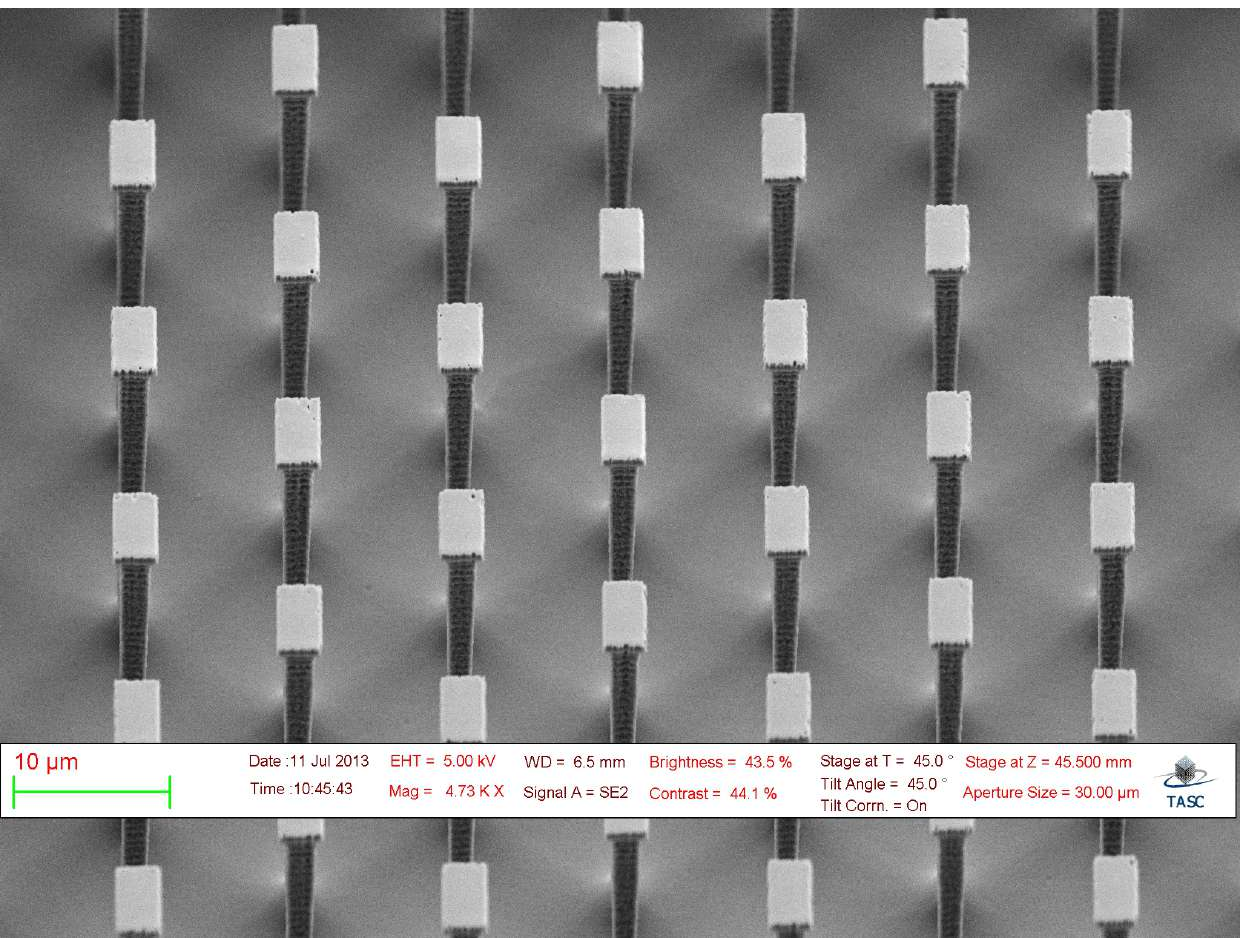}
     \includegraphics[width=0.11\textwidth]{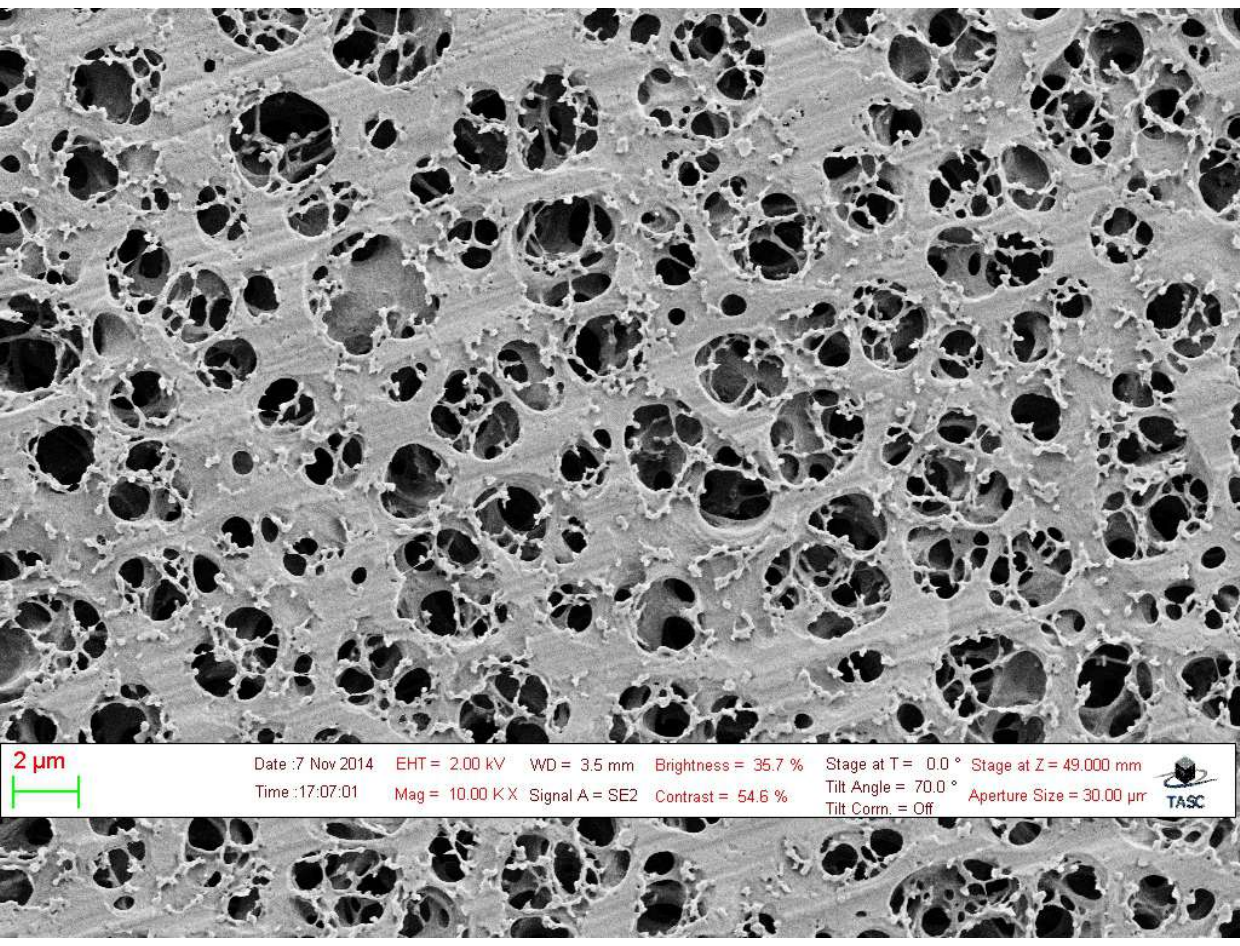}
     \includegraphics[width=0.11\textwidth]{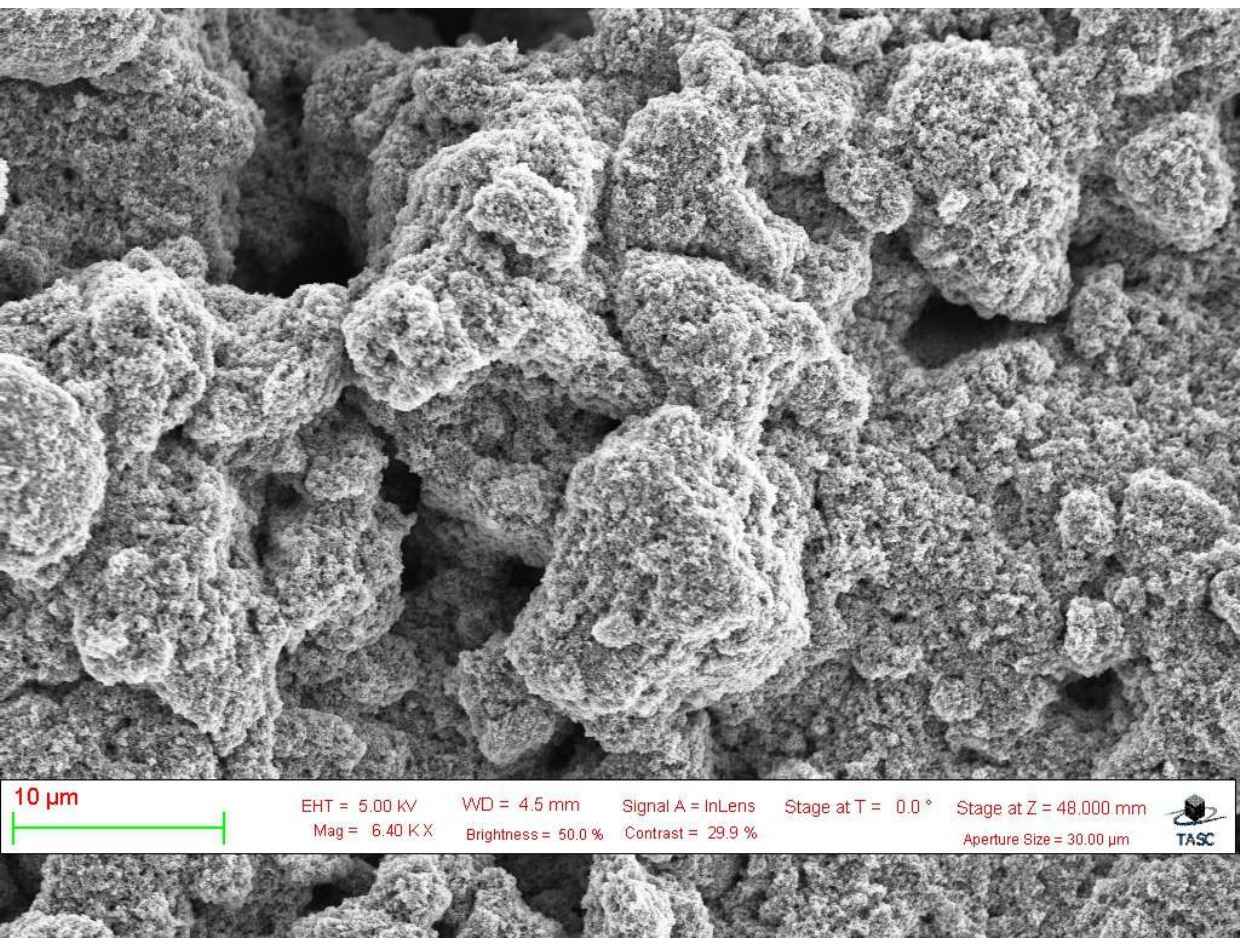}
     }
          \vspace{-13mm}
     \qquad
     \subfloat{\hspace{-0mm}\includegraphics[width=0.11\textwidth]{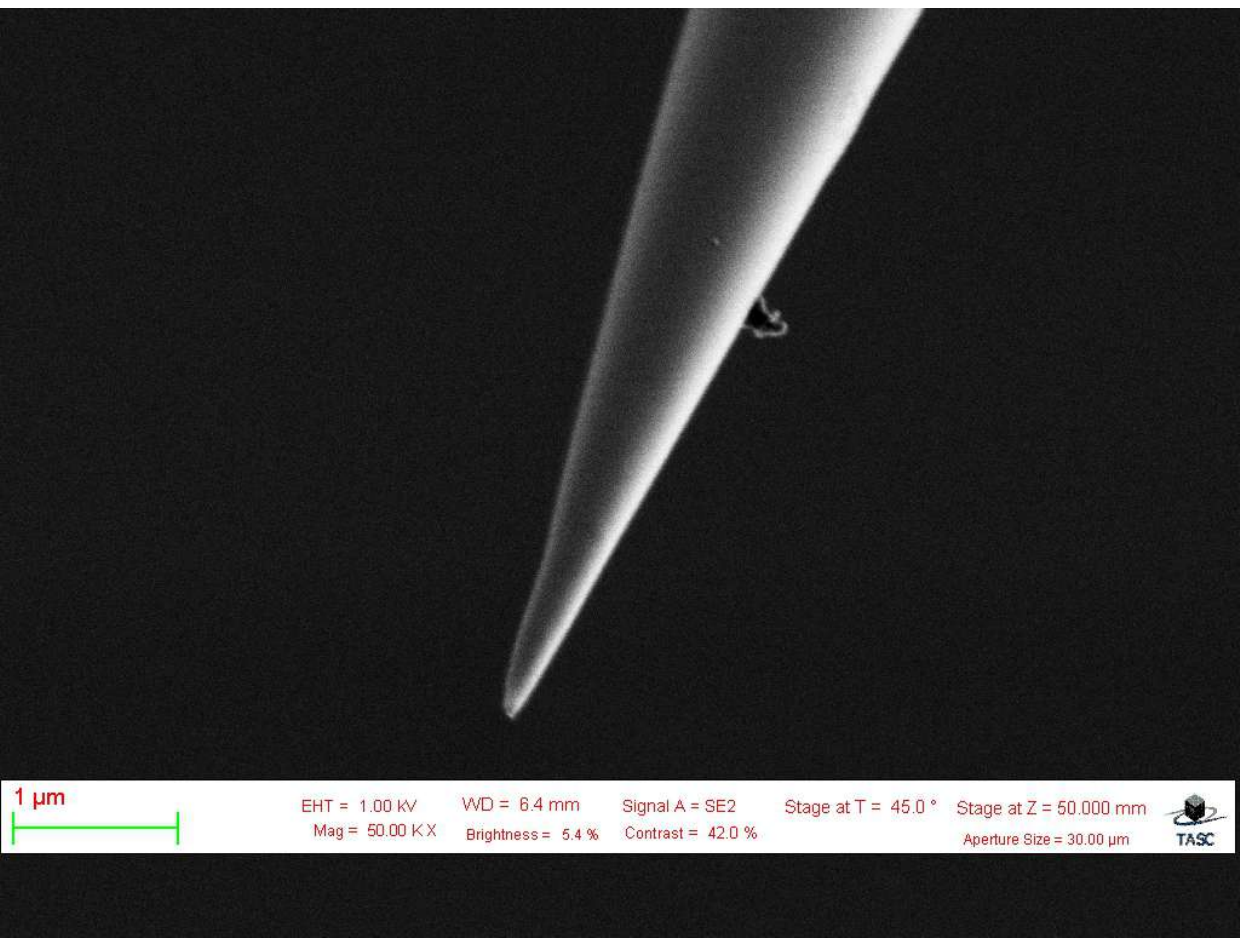}
     }
     \vspace{-10mm}
     \caption{The figure showcases nanomaterials from the SEM dataset\cite{aversa2018first}. In the first, second and third rows (from left to right), we have: \textit{biological, fibers, films}, \textit{MEMS, nanowires, particles}, and \textit{patterned surface, porous sponges, powder}, respectively. Meanwhile, the last row displays: \textit{tips}.}
      \vspace{0mm}
     \label{fig:illustrationpics}
\vspace{-4mm}     
\end{figure}

\vspace{-1mm}
\paragraph{Datasets:} Our study utilized the SEM dataset\cite{aversa2018first} to automate nanomaterial identification. The expert-annotated dataset spans 10 distinct categories, representing a range of nanomaterials, such as particles, nanowires, and patterned surfaces, among others. It contains approximately 21,283 electron micrographs. Figure \ref{fig:illustrationpics} provides a representation of the different nanomaterial categories in the SEM dataset. Despite initial findings\cite{modarres2017neural} on a subset, our research was based on the complete dataset. The original dataset curators\cite{aversa2018first}, did not provide predefined splits for training, validation, and testing, so we employed the k-fold cross-validation method. This strategy facilitated a fair and rigorous comparison with popular baseline models. To further validate our proposed framework, we evaluated it on several open-source material benchmark datasets relevant to our study, encompassing diverse applications. This allowed us to demonstrate the efficacy of our framework and its applicability to a broader range than just the SEM dataset.

\vspace{-9.5mm}

\vspace{-1mm}
\begin{table*}[ht!]
\footnotesize
\centering
\setlength{\tabcolsep}{5pt}
\caption{The table compares our proposed method to baseline algorithms, such as vision-based supervised convolutional neural networks (ConvNets), vision transformers (ViTs), and self-supervised learning (SSL) algorithms.}
\label{tab:table2}
\vspace{-3mm}
\begin{tabular}{c|c|c|c|c|c}
\hline
\multicolumn{2}{c|}{\textbf{Algorithms}} & \textbf{Top-1} & \textbf{Top-2} & \textbf{Top-3} & \textbf{Top-5} \\ 
\hline
\multirow{6}{*}{\rotatebox[origin=c]{90}{\textbf{ConvNets}}} & AlexNet\cite{krizhevsky2017imagenet} & 0.493 & 0.582 & 0.673 & 0.793 \\
& DenseNet\cite{huang2017densely} & 0.539 & 0.750 & 0.875 & 0.906 \\
& ResNet\cite{he2016deep} & 0.512 & 0.766 & 0.891 & 0.906 \\
& VGG\cite{simonyan2014very} & 0.517 & 0.644 & 0.717 & 0.779 \\
& GoogleNet\cite{szegedy2015going} & 0.560 & 0.844 & 0.906 & 0.938 \\
& SqueezeNet\cite{iandola2016squeezenet} & 0.436 & 0.469 & 0.609 & 0.656 \\ 
\hline
\multirow{6}{*}{\rotatebox[origin=c]{90}{\textbf{VSL}}} & Barlowtwins\cite{zbontar2021barlow} & 0.138 & 0.250 & 0.328 & 0.453 \\
& SimCLR\cite{chen2020simple} & 0.157 & 0.234 & 0.359 & 0.469 \\
& byol\cite{grill2020bootstrap} & 0.130 & 0.234 & 0.281 & 0.422 \\
& moco\cite{he2020momentum} & 0.158 & 0.188 & 0.250 & 0.438 \\
& nnclr\cite{dwibedi2021little} & 0.144 & 0.266 & 0.313 & 0.531 \\
& simsiam\cite{chen2021exploring} & 0.170 & 0.266 & 0.391 & 0.500 \\ 
\hline
\multirow{24}{*}{\rotatebox[origin=c]{90}{\textbf{ViTs}}} & CCT\cite{hassani2021escaping} & 0.600 & 0.781 & 0.875 & 0.969 \\
& CVT\cite{CVT} & 0.537 & 0.750 & 0.828 & 0.953 \\
& ConViT\cite{ConViT} & 0.582 & 0.734 & 0.828 & 0.938 \\
& ConvVT\cite{CVT} & 0.291 & 0.563 & 0.734 & 0.875 \\
& CrossViT\cite{Crossvit} & 0.466 & 0.719 & 0.828 & 0.938 \\
& PVTC\cite{PVT} & 0.567 & 0.766 & 0.813 & 0.922 \\
& SwinT\cite{SwinT} & 0.675 & 0.766 & 0.891 & 0.938 \\
& VanillaViT\cite{dosovitskiy2020image} & 0.623 & 0.828 & 0.859 & 0.938 \\
& Visformer\cite{visformer} & 0.371 & 0.578 & 0.641 & 0.797 \\
& ATS\cite{fayyaz2021ats} & 0.511 & 0.703 & 0.828 & 0.938 \\
& CaiT\cite{CaiT} & 0.616 & 0.750 & 0.906 & 0.938 \\
& DeepViT\cite{Deepvit} & 0.512 & 0.734 & 0.875 & 0.938 \\
& Dino\cite{Dino} & 0.047 & 0.219 & 0.391 & 0.432 \\
& Distallation\cite{Distillation} & 0.516 & 0.719 & 0.844 & 0.938 \\
& LeViT\cite{Levit} & 0.597 & 0.813 & 0.875 & 0.953 \\
& MA\cite{MA} & 0.192 & 0.288 & 0.350 & 0.459 \\
& NesT\cite{Nest} & 0.636 & 0.828 & 0.891 & 0.953 \\
& PatchMerger\cite{PatchMerger} & 0.549 & 0.719 & 0.859 & 0.922 \\
& PiT\cite{PiT} & 0.520 & 0.703 & 0.828 & 0.953 \\
& RegionViT\cite{Regionvit} & 0.575 & 0.797 & 0.859 & 0.922 \\
& SMIM\cite{SMIM} & 0.163 & 0.297 & 0.453 & 0.609 \\
& T2TViT\cite{T2TViT} & 0.702 & 0.859 & 0.906 & 0.938 \\
& ViT-SD\cite{ViT-SD} & 0.613 & 0.766 & 0.906 & 0.953 \\ \hline
& \texttt{CM-EMRL} & \textbf{0.9161} & \textbf{0.9339} & \textbf{0.9691} & \textbf{0.9719} \\ 
\bottomrule
\end{tabular}
\vspace{-4mm}
\end{table*}

\begin{figure*}[ht!]
\centering
\resizebox{0.725\linewidth}{!}{ 
\hspace*{5mm}\includegraphics[keepaspectratio,height=4.5cm,trim=0.0cm 0.0cm 0cm 0.0cm,clip]{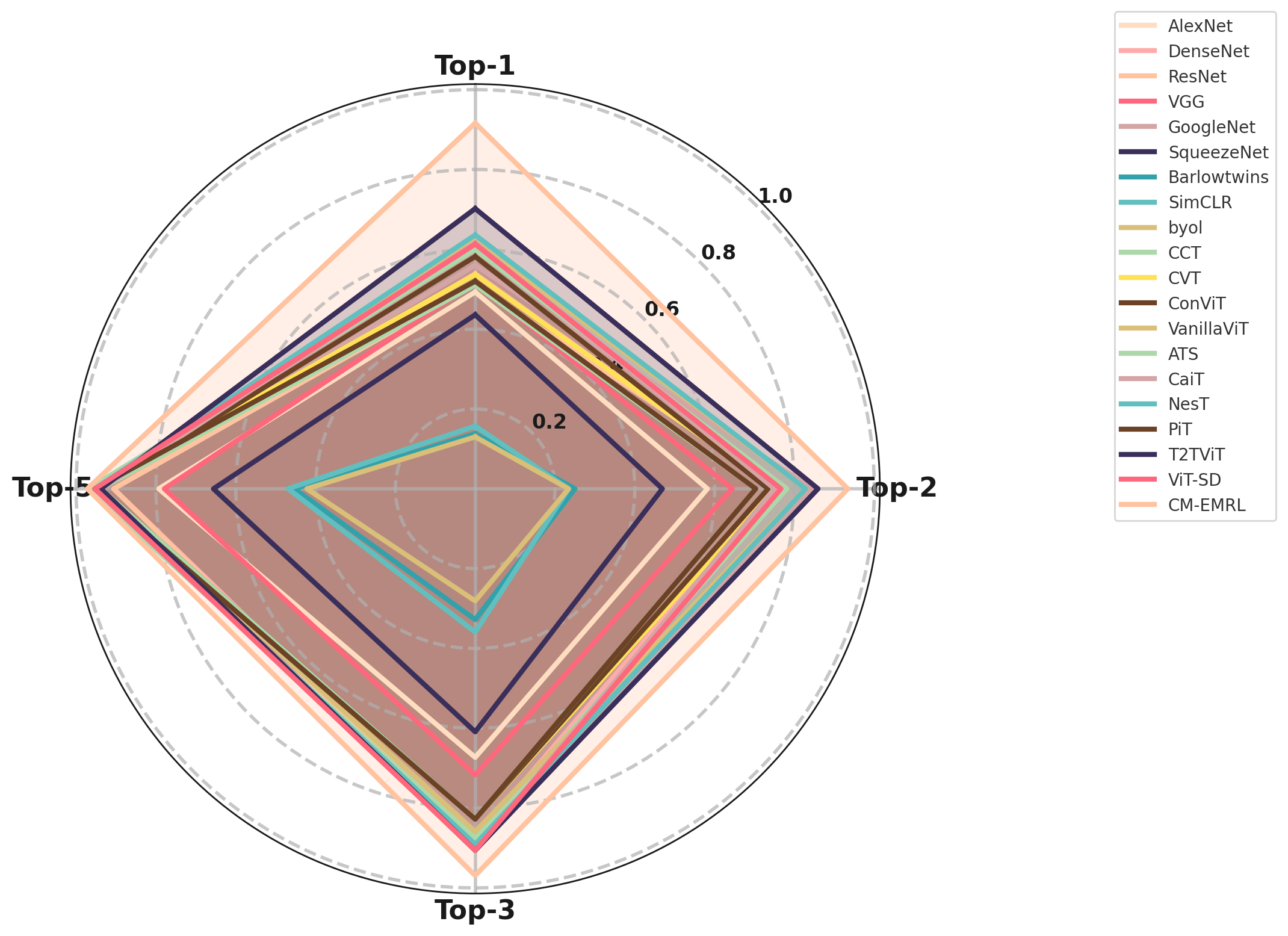} % left, bottom, right, top
}
\vspace{-1mm}
\caption{The figure compares our proposed framework to vision-based supervised convolutional neural networks (ConvNets), vision transformers (ViTs), and self-supervised learning (VSL) algorithms on the SEM dataset \cite{aversa2018first}.}
\label{fig:figure5}
\vspace{0mm}
\end{figure*}

\begin{figure*}[ht!]
\centering
\resizebox{0.75\linewidth}{!}{ 
\hspace*{5mm}\includegraphics[keepaspectratio,height=4.5cm,trim=0.0cm 0.0cm 0cm 0.0cm,clip]{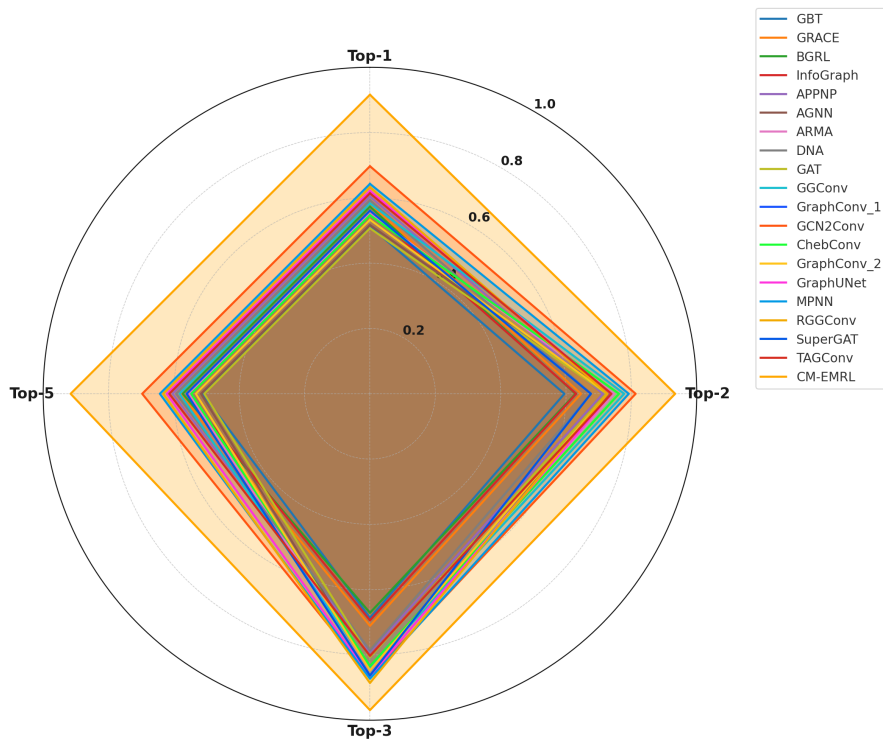} % left, bottom, right, top
}
\vspace{-1mm}
\caption{The figure shows the performance comparison of the proposed framework to supervised GNNs and self-supervised GCL algorithms on the SEM dataset \cite{aversa2018first}.}
\label{fig:figure6}
\vspace{-3mm}
\end{figure*}

\vspace{-2mm}
\paragraph{Results:} To measure the effectiveness of our proposed framework, we conducted an in-depth analysis contrasting it with popular computer vision baseline models. We compared our framework to supervised learning models, notably Convolutional Neural Networks (ConvNets) and Vision Transformers \cite{philvformer, neelayvformer}, and self-supervised approaches like Vision Contrastive Learning \cite{susmelj2020lightly}. The results of this analysis are shown in Table \ref{tab:table2}. To ensure an fair and rigorous comparison, all tests were conducted under uniform settings across different algorithms. We assessed performance using the Top-$N$ accuracy metric, specifically for $N$ values of 1, 2, 3, and 5. Notably, our framework outperformed the best-performing baseline model, T2TViT (\cite{T2TViT}), demonstrating a significant $30.50\%$ improvement in Top-1 accuracy and a modest $3.61\%$ gain in Top-5 accuracy. Table \ref{tab:table3} presents a comparison between our framework and a selection of supervised learning-based baseline models. This includes several GNN architectures \cite{rozemberczki2021pytorch, Fey/Lenssen/2019} as well as Graph Contrastive Learning (GCL) algorithms \cite{Zhu:2021tu}. Impressively, our framework establishes a new state-of-the-art benchmark, outperforming all other baselines on the benchmark dataset\cite{aversa2018first}. Figures \ref{fig:figure5} and \ref{fig:figure6} shows the radar charts corresponding to the results shown in Tables \ref{tab:table2} and \ref{tab:table3}. The underlying hypothesis of our framework is that ViTs can be employed for initial explorations and the generation of baseline results in this context. Zero-shot CoT prompting of LLMs can be leveraged to enhance the initial outcomes of ViTs by utilizing the implicit domain-specific knowledge embedded within the language model's trainable parameters to obtain expressive cross-modal embeddings. On the other hand, few-shot (in-context) learning of LMMs can be utilized to further refine the framework's predictions by providing demonstrations from the training data, potentially leading to a more robust and accurate predictive framework for nanomaterial category prediction. The experimental findings validate this hypothesis and further advancements in the semiconductor industry—a domain where traditional deep learning techniques often underperform due to their lack of a holistic and nuanced approach. Such shortcomings could hinder breakthroughs in the semiconductor industry.

\vspace{-3mm}
\paragraph{Related Work:} The landscape of computer vision has been profoundly influenced by convolutional networks (ConvNets or CNNs). The pioneering LeNet\cite{lecun1998gradient} set the stage for ConvNets, which were subsequently employed in a variety of vision tasks ranging from image classification\cite{krizhevsky2017imagenet} to semantic segmentation\cite{long2015fully}. Over recent years, groundbreaking architectures like ResNet\cite{he2016deep}, MobileNet\cite{howard2017mobilenets}, and NAS\cite{zoph2016neural, yang2020cars} have further refined the capabilities of CNNs. However, the introduction of vision transformers (ViTs)\cite{dosovitskiy2020image, han2022survey, carion2020end, chen2021pre} marked a paradigm shift, leading to the development of numerous enhanced ViT variants. These advances encompass pyramid architectures\cite{SwinT, wang2021pyramid}, local attention mechanisms\cite{han2021transformer, SwinT}, and innovative position encoding methods\cite{wu2021rethinking}. Drawing inspiration from ViTs, the computer vision community has also delved into the potential of Multilayer Perceptrons (MLP) for vision tasks\cite{touvron2022resmlp, tolstikhin2021mlp}. Current vision-based frameworks in the semiconductor manufacturing sector fall short in various aspects, especially when compared to the recently proposed advancements in generative deep learning and multimodal learning. Many existing solutions fail to capitalize on the detailed analysis achievable through the synergy of LLMs, LMMs, and small-scale LMs with electron micrographs. Moreover, the existing frameworks typically analyze electron micrographs (nano images) at a singular modality, through the use of architectures such as ConvNets, ViTs, or MLP-Mixer, missing the opportunities that a multi-modality fusion approach could offer in enhancing classification accuracy. Furthermore, the industry has not fully embraced the utilization of zero-shot CoT LLMs prompting for generating technical descriptions of nanomaterials, which can significantly enhance domain-specific insights essential for nanomaterial identification tasks. Furthermore, the semiconductor manufacturing sector has not fully tapped into the emerging in-context learning capabilities of LMMs with few-shot prompting for predictive nanomaterial analysis, even though these capabilities could significantly enhance the accuracy of nanomaterial predictions. This glaring gap in the integration of image-based and linguistic insights renders current architectures less comprehensive and nuanced, potentially impeding breakthroughs in the semiconductor industry. Instead of relying solely on conventional classification methods, the new framework incorporates both image-based and linguistic insights by leveraging the capabilities of ViTs and LLMs, respectively, as well as the predictive abilities of LMMs. This framework aims to facilitate a more comprehensive and nuanced analysis of electron micrographs, holding significant promise for advancements in the semiconductor industry through automated nanomaterial identification. These advancements highlight the ongoing push for innovation in semiconductor manufacturing, driven by the escalating demand for more powerful and efficient electronic devices.
 
\vspace{-7mm}
\section{Conclusion}
\vspace{-1mm}
To conclude, we conducted the first in-depth study aimed at achieving state-of-the-art performance in nanomaterial characterization. We have introduced an innovative framework that employs ViTs as the foundational layer, further enriched through the multi-modal fusion approach of zero-shot CoT prompting of LLMs and refined with few-shot (in-context) learning of LMMs. Our experiments confirm the superiority of this framework, indicating its transformative potential for semiconductor manufacturing in the age of advanced electronic devices.

\vspace{-5mm}
\bibliography{aaai24}

\vspace{-2mm}
\section{Technical Appendix}

\vspace{-2mm}

\vspace{-5mm}
\begin{table*}[ht!]
\footnotesize
\centering
\setlength{\tabcolsep}{4pt}
\caption{The table shows the performance comparison of the proposed method with supervised GNNs and self-supervised GCL algorithms on the SEM dataset\cite{aversa2018first}.}
\label{tab:table3}
\vspace{-3mm}
\begin{tabular}{c|c|c|c|c|c}
\hline
\multicolumn{2}{c|}{\textbf{Algorithms}} & \textbf{Top-1} & \textbf{Top-2} & \textbf{Top-3} & \textbf{Top-5} \\ 
\hline
\multirow{4}{*}{\rotatebox[origin=c]{90}{\textbf{GSL}}} & GBT\cite{bielak2021graph} & 0.513 & 0.595 & 0.686 & 0.778 \\
& GRACE\cite{zhu2020deep} & 0.581 & 0.646 & 0.711 & 0.773 \\
& BGRL\cite{thakoor2021bootstrapped} & 0.573 & 0.629 & 0.671 & 0.728 \\
& InfoGraph\cite{sun2019infograph} & 0.560 & 0.631 & 0.694 & 0.756 \\
\hline
\multirow{15}{*}{\rotatebox[origin=c]{90}{\textbf{GNN}}} & APPNP\cite{klicpera2018predict} & 0.604 & 0.713 & 0.792 & 0.823 \\
& AGNN\cite{thekumparampil2018attention} & 0.517 & 0.733 & 0.841 & 0.943 \\
& ARMA\cite{bianchi2021graph} & 0.553 & 0.747 & 0.848 & 0.925 \\
& DNA\cite{fey2019just} & 0.593 & 0.677 & 0.786 & 0.891 \\
& GAT\cite{velivckovic2017graph} & 0.507 & 0.724 & 0.807 & 0.914 \\
& GGConv\cite{li2015gated} & 0.583 & 0.778 & 0.841 & 0.944 \\
& GraphConv\cite{morris2019weisfeiler} & 0.533 & 0.727 & 0.847 & 0.961 \\
& GCN2Conv\cite{chen} & 0.697 & 0.813 & 0.867 & 0.945 \\
& ChebConv\cite{defferrard2016convolutional} & 0.547 & 0.762 & 0.834 & 0.896 \\
& GraphConv\cite{morris2019weisfeiler} & 0.533 & 0.727 & 0.847 & 0.961 \\
& GraphUNet\cite{gao2019graph} & 0.622 & 0.738 & 0.866 & 0.912 \\
& MPNN\cite{gilmer2017neural} & 0.643 & 0.792 & 0.873 & 0.959 \\
& RGGConv\cite{bresson2017residual} & 0.633 & 0.727 & 0.886 & 0.928 \\
& SuperGAT\cite{kim2022find} & 0.561 & 0.676 & 0.863 & 0.935 \\
& TAGConv\cite{du2017topology} & 0.614 & 0.739 & 0.803 & 0.946 \\
\hline
& \texttt{CM-EMRL} & \textbf{0.9161} & \textbf{0.9339} & \textbf{0.9691} & \textbf{0.9719} \\ 
\bottomrule
\end{tabular}
\vspace{-2mm}
\end{table*}

\vspace{-2mm}

\vspace{8mm}
\subsection{Ablation Study}
\vspace{0mm}
Figure \ref{fig:figure3} illustrates an overview of the framework. The proposed framework involves four components: (a) The first component, the electron micrograph encoder, takes an input image and divides it into smaller patches. These patches are transformed into tokens, enriched with positional embeddings for spatial information, and a classification token is added as a separate token to represent the overall image content. The resulting augmented token sequence is fed into a ViT model to generate an embedding that represents the entire image. (b) Next, the zero-shot CoT LLMs prompting technique uses cloud services to access LLMs for generating detailed descriptions about various aspects of nanomaterials. Structured prompts guide the LLMs in generating in-depth descriptions on topics ranging from the fundamental properties of nanomaterials to their practical applications. Subsequently, we fine-tune smaller LMs on the descriptions generated by the LLMs to obtain context-aware token embeddings. We then perform sum-pooling attention mechanism to obtain contextualized text-level embeddings, which capture the core knowledge in the generated texts. (c) Finally, the cross-modal alignment employs the scaled dot-product attention mechanism to match the image embeddings with the corresponding nanomaterial-specific text-level embeddings. This process highlights image features relevant to the textual descriptions, aiding in the identification of text-level embeddings that correspond to the image. (c) Few-shot prompting enables LMMs to quickly adapt to new tasks without traditional fine-tuning on labeled data. Using a small set of input-output pairs, these models learn tasks by drawing from their vast pre-existing knowledge acquired during training on vast, diverse text corpora. In the context of nanomaterial identification, LMMs utilize a handful of image-label pairs from the training data to classify new, unseen images and obtain prediction embeddings. Demonstrations can be selected either randomly or based on their similarity to the (c) Few-shot prompting enables LMMs to quickly adapt to new tasks without traditional fine-tuning on labeled data. Using a small set of input-output pairs, these models learn tasks by drawing from their vast pre-existing knowledge acquired during training on vast, diverse text corpora. In the context of nanomaterial identification, LMMs utilize a handful of image-label pairs from the training data to classify unseen query images and obtain prediction embeddings. Demonstrations can be selected either randomly or based on their similarity to the query image. (d) Finally, the unified attention layer, through the hierarchical multi-head attention mechanism, combines information from the original image embedding, a text-level embedding, and the prediction embedding, optimizing for accuracy in nanomaterial categorization. To evaluate the efficacy of the individual components and validate the design choices for their inclusion in the framework, we conducted an ablation study. In this study, we selectively disabled specific components to create various ablated variants, which were then evaluated using the SEM dataset\cite{aversa2018first} on nanomaterial identification. Compared to our proposed original framework, which serves as the baseline, the ablated variants exhibited a notable decline in performance, underscoring the importance of the components that were disabled. The ablation study results support the hypothesis that each component is crucial for the framework’s peak performance in nanomaterial identification. Ablated variants excluding the zero-shot CoT LLMs prompting with cross-modal alignment, few-shot prompting with LMMs, and unified attention methods are labeled as proposed framework ``w/o LLMs", ``w/o LMMs", and ``w/o MHA"; ``w/o" is shorthand for ``without". In the case of ``w/o MHA'', we concatenate the cross-domain embeddings and then transform them through a linear layer to predict the label. The findings from the ablation study are presented in Table \ref{tab:table4}. The `w/o LLMs' variant exhibits a significant decrease in performance compared to the baseline, with a noteworthy drop of $19.18\%$  in \textbf{Avg-Precision}, underscoring the crucial role of zero-shot CoT LLMs in prompting. This technique extracts detailed nanomaterial technical descriptions from closed-source large language models (LLMs), then fine-tunes smaller language models on these descriptions to produce context-aware token embeddings. It subsequently employs weighted sum-pooling mechanism to derive text-level embeddings. 

\vspace{-4mm}
\begin{table}[ht!]
\footnotesize
\centering
\setlength{\tabcolsep}{3.5pt}
\vspace{-3mm}
\resizebox{0.45\textwidth}{!}{%
\subfloat{%
\begin{tabular}{cc|c|c|c|c|c|c|cc}
\hline
\multicolumn{2}{c|}{\textbf{Algorithms}}                                     &  \textbf{Avg-Precision} & \textbf{Avg-Recall} & \textbf{Avg-F1 Score}  \\ \toprule
\multicolumn{1}{c}{\multirow{4}{*}{\rotatebox[origin=c]{90}{\textbf{}}}} & \texttt{CM-EMRL}   & \textbf{0.9157} & \textbf{0.9125} & \textbf{0.9120} \\ \hline 
\multicolumn{1}{c}{}                                          &    w/o LLMs   & 0.7401 	& 0.7936 	&  0.8141 \\
\multicolumn{1}{c}{}                                          &    w/o LMMs    &  0.8160 	& 0.7804 	&  0.8053 \\ 
\multicolumn{1}{c}{}                                          &    w/o MHA    & 0.6937 	& 0.7180 	& 0.7159 \\ \hline
\end{tabular}}}
\vspace{-2mm}
\caption{In our ablation study, we methodically deactivate individual components to evaluate their unique contributions. The aim is to gauge how specific components influence the framework's overall performance. The experiments underscore the importance of each disabled component, as evidenced by the marked drop in performance metrics when compared to the baseline. These results validate our hypothesis that the joint optimization of the zero-shot CoT LLMs prompting with cross-modal alignment, few-shot prompting with LMMs, and unified attention methods enhances the overall framework's efficacy.}
\label{tab:table4}
\vspace{-4mm}
\end{table}

\newpage
Cross-modal alignment correlates these text embeddings with image embeddings via scaled dot-product attention, thereby synchronizing and matching information across different types of data modalities in a unified representation space. Similarly, the `w/o LMMs' variant performs notably worse than the baseline, exhibiting a drop of $10.88\%$ in \textbf{Avg-Precision}. This underscores the significance of few-shot prompting to predict the label of query image, which leverages LMMs to rapidly adapt to new tasks with limited demonstrations (image-label pairs) and generate prediction embeddings. In addition, the `w/o MHA' variant showed a significant performance deterioration compared to the baseline, marked by a $24.24\%$ drop in \textbf{Avg-Precision}. This deterioration can largely be attributed to the overly simplified linear operator in the output layer. This further emphasizes the importance of the unified attention layer, which amalgamates insights from image, text, and prediction embeddings using a hierarchical multi-head attention mechanism to optimize nanomaterial categorization accuracy. Similarly, in comparison to the baseline, the `w/o LLMs' variant exhibits a 13.03$\%$ decrease in \textbf{Avg-Recall}, the `w/o LMMs' variant shows a 14.48$\%$ reduction, and the `w/o MHA' variant experiences a significant drop of 21.31$\%$ in the same metric. Our ablation study highlighted the significant contributions of each component within our framework. When individual components were omitted, we observed a consistent drop in performance. These findings validate the integral role that each component plays in achieving optimal performance of the holistic framework. In summary, each component serves a specific purpose, and their integration ensures a comprehensive approach to nanomaterial categorization. Their inclusion is justified by the need for detailed analysis, adaptability, and holistic consideration of both visual and textual data in categorizing nanomaterials.

\vspace{-1mm}
\subsection{Empirical Insights into Nanomaterial Classification}
\begin{figure*}[ht!]
\centering
\resizebox{0.585\linewidth}{!}{ 
\hspace*{-0mm}\includegraphics[keepaspectratio,height=4.5cm,trim=0.0cm 0.0cm 0cm 0.0cm,clip]{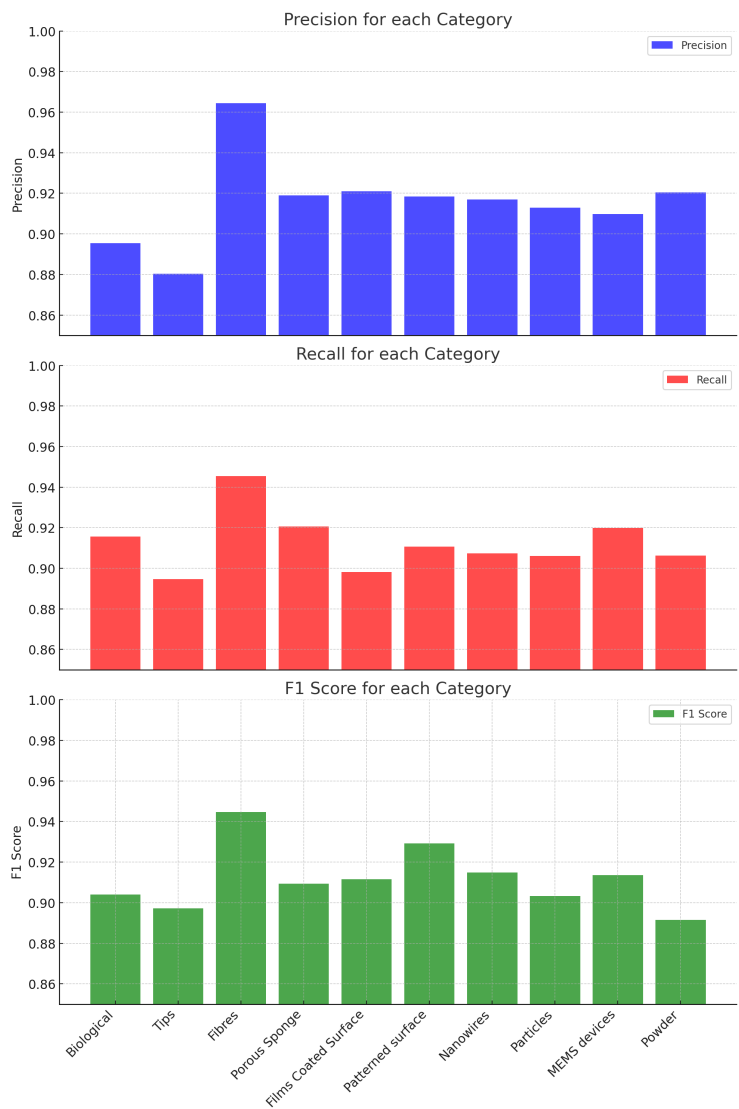} % left, bottom, right, top
}
\vspace{-3mm}
\caption{The bar plots illustrate the precision, recall, and F1-score metrics for the framework's performance across various nanomaterial categories, as depicted by electron micrographs within the SEM dataset. Each category is represented with a distinct light shade for clarity.}
\label{fig:figure4}
\vspace{-4mm}
\end{figure*}

\vspace{-1mm}
We conducted comprehensive experiments to assess our proposed framework's effectiveness in classifying electron micrographs of various nanomaterials, from simple to intricate patterns. Our primary goal is to highlight the framework's classification capabilities. Nanomaterials exhibit a diverse range of patterns due to differences in composition, morphology, surface properties, crystallinity, and synthesis methods. These patterns, as captured in electron micrographs, reflect the materials' unique properties, structures, and potential applications. They can depict anything from isolated nanoparticles to complex aggregations, and from crystalline to amorphous structures. Accurate interpretation of these patterns is vital for understanding and leveraging each nanomaterial category's distinctive qualities. Electron micrographs provide detailed nanoscale insights, revealing structural and morphological features essential for various applications in materials science and related fields. As demonstrated by the experimental results displayed in Figure \ref{fig:figure4}, our framework can effectively generalize across diverse nanomaterials, even those with complex patterns. The figure presents bar plots of the framework’s performance on multiple metrics, color-coded by category. We evaluated its performance on the SEM dataset\cite{aversa2018first} using standard metrics such as precision (P in $\%$), recall (R in $\%$), and F1-score (F1 in $\%$). These findings underscore the framework's effectiveness and emphasize its relevance in materials science and nanotechnology. We employ a comprehensive multi-metric evaluation for robust comparison with baseline models, anchored by a confusion matrix that details our framework’s performance in classifying electron micrographs across nanomaterial categories. The matrix includes key metrics for multi-class classification: True Positives (TP), which are micrographs correctly identified as belonging to a category; False Negatives (FN), which are micrographs that belong to a category but are incorrectly classified as not belonging; True Negatives (TN), which are micrographs correctly identified as not belonging to a category; and False Positives (FP), which are micrographs incorrectly identified as belonging to a category. From these metrics, we calculate precision (the ratio of correctly classified micrographs to the total classified as belonging to a category, TP / (TP + FP)), recall (the proportion of actual micrographs from a category that were correctly classified, TP / (TP + FN)), and the F1-score, which is the harmonic mean of precision and recall, providing an overall measure of effectiveness in micrograph categorization. Thus, our multi-metric approach, anchored by the confusion matrix and its derived metrics, ensures a rigorous evaluation of our framework. This approach facilitates a detailed understanding of our model's effectiveness in categorizing electron micrographs across various nanomaterial categories. It is crucial to highlight that the SEM dataset exhibits significant class imbalance. Notably, our framework achieves higher classification scores for nanomaterial categories with a substantial number of labeled instances, outperforming its performance on categories with fewer instances. The notable success with fewer labeled instances can be attributed to our proposed framework's reduced reliance on nanomaterial-specific relational inductive biases, which distinguishes it from conventional methods. In conclusion, our extended experiments have bolstered our confidence in the framework's ability to generalize and accurately categorize various nanomaterials using electron micrographs. We believe that these advancements will greatly benefit the wider community by accelerating materials characterization and related research. 

\vspace{-2mm}

\vspace{0mm}
\subsection{Experimental Setup}
\vspace{0mm}
The SEM dataset~\cite{aversa2018first} comprises electron micrographs with a resolution of $1024 \times 768 \times 3$ pixels. For our analysis, we resized these to $224 \times 224 \times 3$ pixels. During preprocessing, we normalized the images, adjusting their mean and covariance to a consistent value of 0.5 across all channels, ensuring values between [-1, 1]. These downscaled and normalized images were segmented into distinct, non-overlapping patches, treated as a sequence of tokens with a $32$-pixel resolution. Both the patch dimension ($d_{\text{pos}}$) and the position embedding dimension ($d$) were set to $64$. We employed a 10-fold cross-validation for evaluation, training for $50$ epochs with an initial learning rate of $10^{-3}$ and a batch size of $48$. The unifying (cross-modal) attention layer had the number of attention heads (H) set to $4$ and the key/query/value dimensionality ($d_{h}$) to $16$. To optimize the \texttt{CM-EMRL} framework's performance, we used early stopping on the validation set and a learning rate scheduler, decreasing the rate by 50$\%$ if the validation loss did not decrease for five epochs. The Adam optimization algorithm~\cite{kingma2014adam} was used to fine-tune the framework's parameters. Our proposed framework seeks to enhance multi-class classification precision by combining the strengths of large multimodal models (LMMs) such as GPT-4V, large language models (LLMs) like GPT-4, and smaller language models (LMs). The methodology utilizes LLMs to generate detailed technical descriptions for nanomaterials, capturing crucial linguistic insights critical for nanomaterial identification. We access LLMs, such as GPT-4, through a Language Model as a Service (LaMaaS) platform using text-based APIs, with GPT-4's maximum token sequence output being 4096. GPT-4 with Vision (GPT-4V) extends GPT-4 by adding visual processing capabilities, allowing it to analyze image inputs alongside text. While GPT-4 is limited to text processing, GPT-4V can handle both text and image inputs, enabling applications like visual question answering. We employ few-shot prompting with GPT-4V, using a small number of demonstrations (image-label pairs included in the prompts) to guide its understanding and to predict the label of a query image. To optimize resource utilization, we trained our deep learning models, which are built upon the PyTorch framework, on two V100 GPUs, each with 8 GB of GPU memory. Given the significant computational demands of prompting Large Language Models (LLMs) and Large Multimodal Models (LMMs), we repeated each experiment twice and reported the average results.

\vspace{-3mm}
\paragraph{Baseline Algorithms:} Our baseline methods are organized into four main categories. First, we employ Graph Neural Networks (GNNs) for the supervised multi-class classification of vision graphs \cite{rozemberczki2021pytorch, Fey/Lenssen/2019}. Second, we utilize Graph Contrastive Learning (GCL) methods \cite{Zhu:2021tu}, which involve creating multiple correlated graph views through stochastic augmentations of the input graph data to maximize mutual information between these views. GCL methods aim to enhance the similarities between positive graph views and reduce dissimilarities between negative graph views sampled from different images. Typically, these methods use the Graph Attention Network (GAT) \cite{velivckovic2017graph} to learn node-level embeddings, with graph-level embeddings obtained via sum-pooling of node embeddings. For classification, we apply supervised learning using the Random Forest (RF) algorithm, which utilizes these embeddings to predict nanomaterial categories. We then evaluate the efficiency of these unsupervised embeddings by assessing the RF model's accuracy with holdout data. Third, we employ supervised learning with Convolutional Neural Networks (ConvNets) for the classification of electron micrographs \cite{philvformer, neelayvformer}. Lastly, Vision Transformers (ViTs) are used for supervised classification, utilizing sequences of image patches from electron micrographs as input \cite{philvformer, neelayvformer}, while Vision Contrastive Learning (VCL) techniques \cite{susmelj2020lightly} are self-supervised algorithms designed for contrastive learning in computer vision, utilizing the ResNet architecture for feature extraction. As for our data representation, we construct vision graphs from electron micrographs using the Top-K nearest neighbor search, where patches are treated as nodes, and edges represent pairwise associations between semantically similar neighboring nodes. We employ a 32-pixel patch size and choose K=5 for simplicity to avoid multi-scale vision graphs with varying patch resolutions.

\vspace{-3mm}
\paragraph{Hyperparameter Studies} We carried out an extensive hyperparameter tuning for our framework, specifically examining the embedding dimension ($d$) and batch size ($b$). We evaluated $d$ values from the set $\{32, 64, 128, 256\}$ and $b$ values from the set $\{32, 48, 64, 96\}$. Using the random-search approach, we measured performance based on Top-1 classification accuracy on the validation set. This thorough analysis determined the optimal settings for our framework to be $d=64$ and $b=48$, with a corresponding Top-1 accuracy of 0.9161.

\vspace{-2mm}
\begin{table}[ht!]
\centering
\setlength{\tabcolsep}{2pt}
\resizebox{0.325\textwidth}{!}{
\begin{tabular}{@{}c|c|c|c|cc@{}}
\hline
(d, b) & (32, 48) & (64, 48) & (128, 48) & (256, 48) \\
\hline
Accuracy & 0.8997 & 0.9161 & 0.8996 & 0.9061 \\
\hline
\end{tabular}
}
\vspace{-2mm}
\caption{Experimental findings of the hyperparameter study: Set 1.}
\label{table:Hs_set1}
\vspace{-4mm}
\end{table}

\vspace{-4mm}
\begin{table}[ht!]
\centering
\setlength{\tabcolsep}{2pt}
\resizebox{0.3\textwidth}{!}{
\begin{tabular}{@{}c|c|c|c|cc@{}}
\hline
(d, b) & (64,32) & (64, 48) & (64, 64) & (64,96) \\
\hline
Accuracy & 0.9017 & 0.9161 & 0.9036 & 0.9023 \\
\hline
\end{tabular}
}
\vspace{-2mm}
\caption{Experimental findings of the hyperparameter study: Set 2.}
\label{table:Hs_set2}
\vspace{-4mm}
\end{table}

\vspace{-3mm}
\subsection{Benchmarking with open-source material datasets}
\begin{figure}[htbp]
    \centering
    \includegraphics[scale=.4]{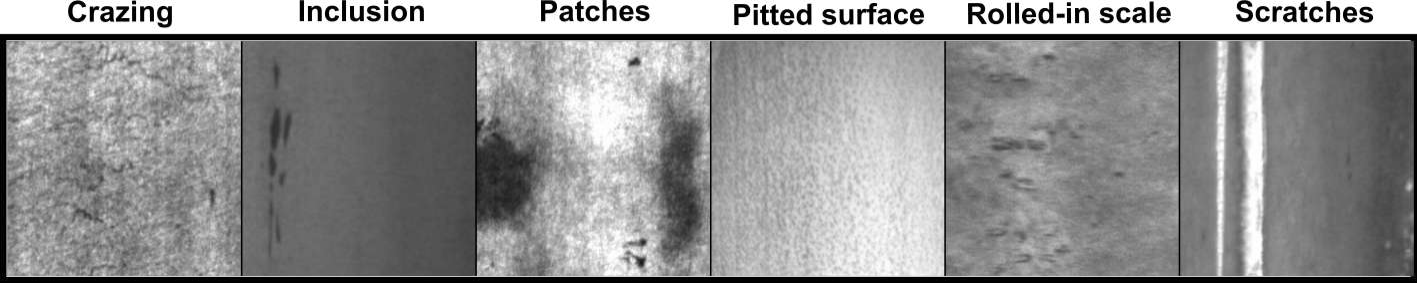}
    \vspace{-5mm}
    \caption{The NEU-SDD dataset comprises six categories of defects in hot-rolled steel strips, as detailed in reference \ref{note1}(\cite{deshpande2020one}).}
    \label{fig:NUE}
     \vspace{-1mm}
\end{figure}

\begin{figure}[htbp]
    \centering
    \includegraphics[scale=0.45]{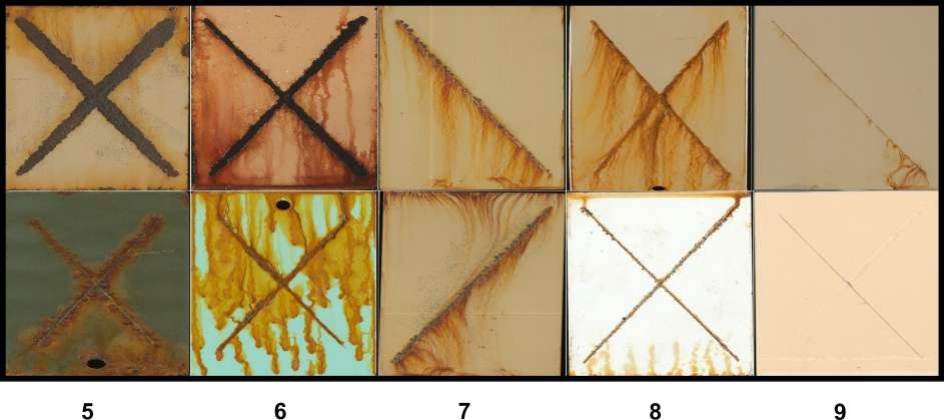}
    \vspace{-2mm}
    \caption{The CMI dataset comprises electron micrographs categorized into five distinct corrosion ratings, as detailed in \ref{note2}. A rating of 9 indicates the least corrosion, while a rating of 5 indicates the most.}
    \label{fig:corrosion}
    \vspace{-5mm}
\end{figure}

\begin{figure}[htbp]
    \centering
    \includegraphics[scale=0.475]{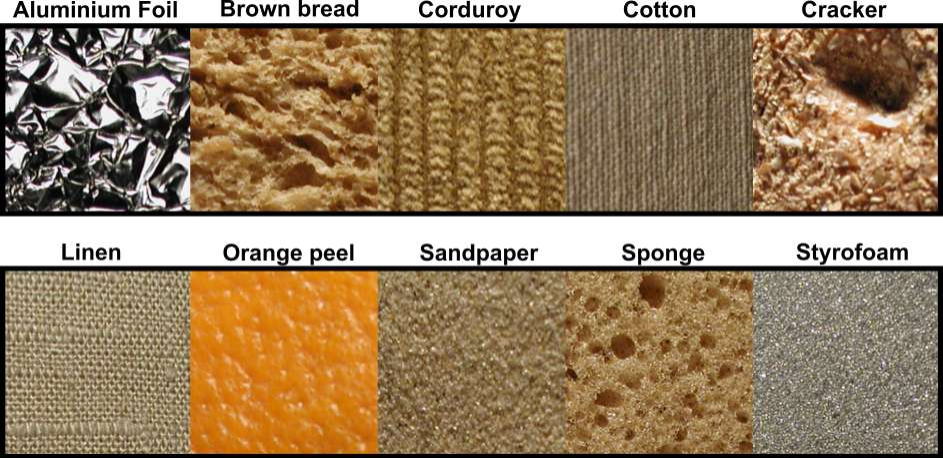}
    \vspace{-2mm}
    \caption{The KTH-TIPS dataset consists of electron micrographs of ten unique materials, which are elaborated upon in reference \ref{note3}.}
    \label{fig:KTH}
     \vspace{-2mm}
\end{figure}

\vspace{1mm}
\begin{itemize}
    \item The NEU-SDD dataset\footnote{Datasource: \url{http://faculty.neu.edu.cn/yunhyan/NEU_surface_defect_database.html}\label{note1}}
(\cite{deshpande2020one}) comprises 1,800 grayscale images captured through electron microscopy, showcasing various surface imperfections in hot-rolled steel strips. This comprehensive dataset is categorized into six types of surface defects—\textit{pitted surfaces, scratches, rolled-in scale, crazing, patches, and inclusions}, with each category represented by 300 images standardized to a resolution of 200$\times$200 pixels. Illustrative examples from these defect categories are presented in Figure \ref{fig:NUE}. Furthermore, we conducted a detailed comparative study using a range of established algorithms to evaluate the proposed method's performance, with a particular focus on multi-class classification tasks for identifying these surface defects.
    \vspace{0mm}      
    \item The CMI dataset\footnote{\url{https://arl.wpi.edu/corrosion_dataset}\label{note2}}
encompasses 600 high-definition electron microscope images that display varying stages of metal panel deterioration. Corrosion experts have meticulously labeled these images following the ASTM-D1654 standard criteria, which employ a scale ranging from 5 to 9 to denote the severity or extent of corrosion. The dataset provides 120 unique images for each level of the corrosion grade defined by this scale, with each image offering a detailed resolution of 512×512 pixels. Figure \ref{fig:corrosion} showcases a set of example images for each corrosion category. We evaluate the effectiveness of our proposed multiclass classification method by benchmarking its ability to accurately assign corrosion grades against a range of established algorithms.  
     \vspace{0mm}          
    \item The KTH-TIPS dataset\footnote{\url{https://www.csc.kth.se/cvap/databases/kth-tips/index.html}\label{note3}}
is a detailed collection of 810 electron micrograph images, each measuring 200$\times$200 pixels, and featuring ten different types of materials. The dataset is notably diverse, including textures such as \textit{sponge, orange peel, styrofoam, cotton, cracker, linen, brown bread, sandpaper, crumpled aluminum foil, and corduroy}, captured under various conditions of lighting, orientation, and scale. Figure \ref{fig:KTH} displays a representative set of these images, providing a visual overview of the dataset's range.  To assess the effectiveness of our proposed technique, we conducted a thorough comparative study, benchmarking it against established algorithms for classifying the various textures or materials represented in the KTH-TIPS dataset.    
\end{itemize}

\vspace{-3mm}

\vspace{-3mm}
Table \ref{tab:table1aux} displays CoT prompts that focus on various types of surface defects. Each prompt is tailored to initiate a comprehensive exploration of the defects characteristics, causes, and impacts. The structured prompts ensure a detailed and systematic exploration, providing clarity and comprehensive coverage. Consequently, this approach yields precise information and a nuanced understanding of the defects implications. Similarly, Table \ref{tab:table2aux} presents a series of CoT prompts designed for an in-depth examination of metal panel corrosion. These prompts guide the research towards a nuanced exploration of different corrosion grades and their effects on metal panels. This structured approach facilitates the study, analysis, and practical application of findings, such as the automatic rating of corrosion in metal panels.

\vspace{-4mm}

\vspace{-3mm}

\vspace{-2mm}
The notion that a single, universal CoT prompt can address all tasks with GPT-4 is inaccurate when it comes to generating technical descriptions of nanomaterials as metadata. Specialized prompts tailored to specific needs are crucial for effectively utilizing AI in the analysis of nanomaterials across different diverse datasets. Customized prompting is essential in AI, as demonstrated by the use of GPT-4 to generate prompts(questions) specific to nanomaterial categories in various datasets, as opposed to relying on a single universal CoT prompt for all across various datasets.

\vspace{-2mm}
\begin{table}[ht!]
\centering
\caption{Customized CoT prompts for detailed analysis of material surface defects.}
\label{tab:table1aux}
\vspace{-5mm}
\begin{tcolorbox}[colback=white!5!white,colframe=black!75!black]
\vspace{-2mm}
\textbf{Prompt 1:} Overview: Briefly describe the specific material surface defect and its impact on material performance. \textbf{Prompt 2:} Characteristics: Define the defect and its identifying features. \textbf{Prompt 3:} Formation: Discuss the formation mechanisms of the defect. \textbf{Prompt 4:} Detection: List the primary techniques for defect detection. 
\textbf{Prompt 5:} Effects: Explain the defect's effects on material properties. \textbf{Prompt 6:} Mitigation: Outline strategies to mitigate the defect. 
\textbf{Prompt 7:} Engineering: Describe surface engineering techniques to control the defect. \textbf{Prompt 8:} Case Studies: Provide examples where defect management improved material use. 
\vspace{-2mm}
\end{tcolorbox}
\end{table}

\vspace{-4mm}
Table \ref{tab:table3aux} presents prompts designed to thoroughly cover a wide range of information exploring the characteristics and applications of various materials and providing detailed technical descriptions of these materials. Table \ref{tab:auxexp} offers a detailed juxtaposition of the efficacy of our method against several standard approaches, with evaluations conducted on multiple datasets. The outcomes of these experiments reveal that our approach consistently attains unparalleled results, highlighting its effectiveness and reliability.

\vspace{-2mm}
\begin{table}[ht!]
\centering
\caption{Customized CoT prompts for detailed analysis of material panel corrosion.}
\label{tab:table2aux}
\vspace{-5mm}
\begin{tcolorbox}[colback=white!5!white,colframe=black!75!black]
\textbf{Prompt 1:} \emph{Corrosion Grading Overview:} Summarize the numerical corrosion grade system. \textbf{Prompt 2:} \emph{Grade Characteristics:} Detail key features of each corrosion grade. \textbf{Prompt 3:} \emph{Influencing Factors:} Discuss factors that influence corrosion grading. \textbf{Prompt 4:} \emph{Deterioration Mechanisms:} Describe deterioration mechanisms by grade. \textbf{Prompt 5:} \emph{Property Impact:} Examine corrosion's impact on metal properties. \textbf{Prompt 6:} \emph{Mitigation Strategies:} Outline preventive measures for each grade. \textbf{Prompt 7:} \emph{Grade Progression Analysis:} Analyze corrosion grade progression over time. \textbf{Prompt 8:} \emph{Rehabilitation Approaches:} Guide on repair or replace decisions by grade. \textbf{Prompt 9:} \emph{Corrosion Management Case Studies:} Present case studies on corrosion management. \textbf{Prompt 10:} \emph{Economic and Safety Considerations:} Discuss economic and safety implications.
\end{tcolorbox}
\vspace{-4mm}
\end{table}

\vspace{-4mm}
\begin{table}[ht]
\centering
\caption{Customized prompts for detailed material analysis.}
\label{tab:table3aux}
\vspace{-5mm}
\begin{tcolorbox}[colback=white!5!white,colframe=black!75!black]
\textbf{Prompt 1:} Contextual Overview: Introduce the material's origin, common use, and relevance. \textbf{Prompt 2:} Properties: Discuss the material's physical and chemical characteristics. \textbf{Prompt 3:} Production: Outline the processes of material preparation or manufacturing. \textbf{Prompt 4:} Structure: Examine the material's structural features and their implications. \textbf{Prompt 5:} Modification: Describe possible modifications to enhance the material's properties. \textbf{Prompt 6:} Longevity: Analyze the material's durability, degradation, and environmental impact. \textbf{Prompt 7:} Applications: Explore diverse applications and uses of the material. \textbf{Prompt 8:} Economic Impact: Reflect on the material's economic significance and societal influence. \textbf{Prompt 9:} Safety: Address health and safety considerations related to the material. 
\textbf{Prompt 10:} Future Outlook: Speculate on potential future developments and research directions.
\vspace{-2mm}
\end{tcolorbox}
\vspace{-5mm}
\end{table}

\vspace{-6mm}
\begin{table}[ht!]
\footnotesize
\centering
\resizebox{0.4\textwidth}{!}{%
\subfloat{%
\setlength{\tabcolsep}{3pt}
\begin{tabular}{cc|cccc}
\hline
\multicolumn{2}{c|}{\textbf{Algorithms}}                                       & \textbf{NEU-SDD} & \textbf{CMI} & \textbf{KTH-TIPS}  \\ \hline
\multicolumn{1}{c|}{\multirow{4}{*}{\rotatebox[origin=c]{90}{\textbf{Baselines}}}} & ResNet                   & 0.906	& 0.928	& 0.941 &             \\
\multicolumn{1}{c|}{}                                          & GoogleNet                & 0.936	& 0.928	& 0.929
              \\
\multicolumn{1}{c|}{}                                          & SqueezeNet                & 0.955	& 0.943	& 0.963
              \\ 
\multicolumn{1}{c|}{}                                          & VanillaViT               & 0.962	& 0.968	& 0.972
 \\ 
\hline
\multicolumn{1}{c|}{}                                          & \textbf{MultiFusion-LLM}                  &     \textbf{0.988}              &      \textbf{0.979}             &    \textbf{0.983}               &                     \\ \hline
\end{tabular}}}
\vspace{-2mm}
\caption{The table displays a comparative analysis of the performance of our framework in relation to a number of benchmark algorithms across various datasets.}
\label{tab:auxexp}
\vspace{-9mm}
\end{table}

\clearpage
\onecolumn

\begin{tabularx}{1.25\textwidth}{bss}
\caption{The table presents the question-answer pairs retrieved from the LLM(GPT-4) for natural language queries about MEMS devices.}\label{tab:lmprompts}\\
\hline
\toprule
\textbf{Query prompt}  & \textbf{LLM-retrieved text} \\
\midrule
\endhead
\textcolor{black}{Introduction: Provide an overview of the MEMS and its significance in various fields?} & Introduction: Micro-Electro-Mechanical Systems (MEMS) combine miniaturized mechanical and electromechanical elements that are made using the techniques of microfabrication. When these structures are taken down to the nanoscale, they enter a realm often referred to as Nano-Electro-Mechanical Systems (NEMS). These systems, constructed of nanomaterials, display properties and capabilities that are vastly different from bulk materials.

\vspace{2mm}
Nanomaterials in MEMS: Nanomaterials are materials with individual units or structures sized between 1 and 100 nanometers in at least one dimension. This size gives them unique mechanical, electrical, and optical properties. Examples include carbon nanotubes, graphene, quantum dots, and metallic nanoparticles. In MEMS applications, nanomaterials offer: 
(a) Enhanced Mechanical Properties: High strength-to-weight ratio and improved flexibility. (b) Advanced Electrical Conductivity: Enhanced charge transfer and quantum tunneling properties, (c) Sensitivity: Due to their high surface area-to-volume ratio, they are incredibly sensitive to environmental changes, making them ideal for sensing applications, and (d) Thermal Conductivity: Some nanomaterials exhibit exceptional heat conduction properties.

\vspace{2mm}
Significance across Various Fields: MEMS nanomaterials have made inroads in various fields due to their unique attributes: (a) Medical and Healthcare: Used in implantable devices, drug delivery systems, and advanced diagnostic tools. For example, MEMS sensors can detect minute changes in physiological parameters, (b) Telecommunications: Nanomaterials enhance the performance of micro-sized antennas, resonators, and optical switches, (c) Consumer Electronics: Accelerometers in smartphones, tiny microphones, and miniature cameras all leverage MEMS technology, (d) Aerospace and Defense: Advanced sensors for navigation, pressure sensing in harsh environments, and micro propulsion systems, (e) Energy: Harvesting ambient energy (e.g., vibrations) to power devices. Nanomaterials also find applications in efficient solar cells and batteries, (f) Environmental Monitoring: Nanoscale sensors can detect minute concentrations of pollutants or pathogens in air and water, and (g) Automotive: MEMS sensors help in real-time tire pressure monitoring, optimizing air/fuel mixtures, and safety systems like airbags.

\vspace{2mm}
Conclusion: The category of MEMS nanomaterials has paved the way for groundbreaking advancements in various fields. These tiny yet robust systems are redefining the borders of what's feasible, leading to the miniaturization of devices, enhancement of performance, and new applications previously deemed unattainable. Their influence spans from the phones in our pockets to the aerospace technologies that traverse our skies, making MEMS nanomaterials a central player in modern science and technology.

\\

\midrule

\textcolor{black}{Definition and Structure: Define the nanomaterial category and describe its typical structure at the nanoscale?} & Definition of MEMS:  MEMS, or Micro-Electro-Mechanical Systems, are miniaturized devices that combine both mechanical and electronic components. They are made using microfabrication techniques, originally derived from the semiconductor industry. MEMS devices integrate functionalities ranging from sensing to actuating in systems at the micrometer scale. When these devices or structures are reduced to the nanoscale, they often fall under the category of Nano-Electro-Mechanical Systems (NEMS).

\vspace{2mm}
Structure of MEMS at the Nanoscale:  MEMS structures, when reduced to the nanoscale, can have distinct characteristics due to quantum effects and increased surface-to-volume ratios. A typical MEMS structure at the nanoscale may include: (a) Substrate Layer: Often made of silicon or another semiconductor material. This forms the base of the MEMS device. (b) Thin Film Layers: These can be conductive, insulating, or piezoresistive depending on the MEMS device's function. Materials often include metals, oxides, or polymers. (c) Mechanical Structures: These can be beams, membranes, cantilevers, or other movable elements that interact with the environment, either by moving in response to external stimuli or by causing movement to achieve a specific function. (d) Electrodes: Used to apply electrical signals to the mechanical structures or to read electrical signals resulting from the motion of these structures. (e) Sensing Elements: At the nanoscale, these elements can be extremely sensitive due to their high surface area-to-volume ratio. They can be designed to detect changes in pressure, temperature, chemical concentration, and other parameters. (f) Actuation Elements: These convert energy (often electrical) into mechanical action. Examples at the nanoscale include nano-tweezers or nano-switches, and (g) Interconnects: Tiny pathways that provide electrical connectivity among different components of the MEMS device.

\vspace{2mm}
In conclusion, MEMS devices, when miniaturized to the nanoscale, retain their integrated electronic and mechanical functionalities but can exhibit unique properties and enhanced performance due to the nanoscale effects. The structure involves a combination of mechanical elements, sensing and actuation capabilities, and electronic interconnects, all intricately designed and layered on a substrate. \\

\midrule

\textcolor{black}{Synthesis Methods: Examine different methods employed for synthesizing or fabricating nanomaterials within this category. Discuss both their advantages and 
limitations :} &  Synthesis Methods for Nanomaterials in MEMS: Nanomaterials used in MEMS devices can be fabricated or synthesized using various techniques. These techniques differ in their mechanisms, complexity, and outcomes. Here's an examination of some common methods, along with their advantages and limitations:

\vspace{2mm}
(1) Top-down Fabrication: (a) Description: This involves scaling down bulk materials to the desired nano-dimensions using external forces or tools. Techniques like photolithography, electron beam lithography, and nano-imprint lithography fall under this category. (b) Advantages: Direct integration with existing semiconductor processes. Precise control over dimensions and patterns. (c) Limitations: Physical constraints limit the achievable dimensions. Potential damage or defects in the structure during the process. 

\vspace{2mm}
(2) Bottom-up Fabrication: (a) Description: This method assembles nanomaterials atom-by-atom or molecule-by-molecule. Techniques include chemical vapor deposition (CVD), molecular beam epitaxy (MBE), and atomic layer deposition (ALD). (b) Advantages: Can achieve true nanoscale dimensions. Potential for defect-free structures. Greater control at the atomic or molecular level. (c) Limitations: May require high temperatures or ultra-high vacuum conditions.
Some processes can be slower and more resource-intensive.

\vspace{2mm}
(3) Sol-Gel Synthesis: (a) Description: A wet chemical technique where metal alkoxides undergo hydrolysis and polymerization reactions to form a colloidal solution (sol) that acts as a precursor for an inorganic network (gel). (b) Advantages: Can be performed at relatively low temperatures.
Versatility in synthesizing various metal oxides. Allows doping with other elements for added functionalities. (c) Limitations: Controlled evaporation is crucial; otherwise, cracks can form. Purity of the end-product may be a concern. 

\vspace{2mm}
(4). Chemical Reduction: (a) Description: This involves reducing metal ions in solution to form metal nanoparticles. Common reducing agents include sodium borohydride or hydrazine. (b) Advantages: Simple and cost-effective. Can produce a wide range of metal nanoparticles. (c) Limitations: Size and shape control can be challenging. Some reducing agents can be hazardous.

\vspace{2mm}
(5) Hydrothermal/Solvothermal Synthesis: (a) Description: This involves the crystallization of materials at high temperatures and pressures in water or organic solvents, respectively. (b) Advantages: Can synthesize a wide range of nanostructures. Offers good control over particle size and morphology. (c) Limitations: Requires specialized equipment for high pressures and temperatures. Scale-up can be challenging.

\vspace{2mm}
In conclusion, the choice of synthesis or fabrication method for nanomaterials in MEMS largely depends on the desired properties, dimensions, and applications of the end product. While some methods offer precise control over dimensions, others might be more suited for bulk synthesis. Advancements in these techniques continue, aiming to address their limitations and make the production of nanomaterials more efficient and scalable.

\\

\midrule

\textcolor{black}{Properties: Highlight the unique physical, chemical, and electronic properties exhibited by nanomaterials in this category. Explain how these properties differ from those of bulk materials.} &  Nanomaterials, due to their reduced dimensions, exhibit distinct physical, chemical, and electronic properties that often differ remarkably from their bulk counterparts. Here's a highlight of these unique attributes and a brief comparison with bulk materials:

\vspace{2mm}
(1). Physical Properties: 

\vspace{1mm}
(a) Surface Area-to-Volume Ratio: Nanomaterials have a much larger surface area relative to their volume. This increased ratio can enhance processes that occur at the surface, such as catalysis or adsorption. Bulk vs Nano: In bulk materials, the majority of atoms are inside the material and not on the surface. In nanomaterials, a significant fraction of atoms or molecules reside on the surface, leading to pronounced surface effects. 

\vspace{1mm}
(b) Mechanical Strength: Some nanomaterials, like carbon nanotubes, exhibit significantly higher strength compared to their bulk forms. Bulk vs Nano: The enhanced strength in nanoscale materials can be attributed to fewer defects as compared to bulk materials, resulting in stronger atomic or molecular interactions.

\vspace{2mm}
2. Chemical Properties: 

\vspace{1mm}
(a) Reactivity: Nanomaterials tend to be more reactive due to their increased surface area and the higher number of surface atoms. Bulk vs Nano: Bulk materials, with fewer surface atoms relative to their volume, generally exhibit lower reactivity compared to their nanoscale counterparts.

\vspace{1mm}
(b) Catalytic Activity: Many nanomaterials serve as effective catalysts because of their enhanced surface reactivity. Bulk vs Nano: In bulk catalysts, only the surface atoms or molecules can interact with reactants, but in nanomaterials, a greater fraction of the material can participate in the catalytic process.

\vspace{2mm}
3. Electronic Properties: 

\vspace{1mm}
(a) Quantum Confinement: When the dimensions of materials are reduced to the nanoscale, electrons are confined in a smaller space, leading to discrete energy levels. This can affect electronic and optical properties. Bulk vs Nano: In bulk semiconductors, electrons have a continuous range of energy states, while in quantum dots (a type of nanomaterial), the energy states are quantized. 

\vspace{1mm}
(b) Electrical Conductivity: The conductivity of nanomaterials can be higher due to fewer grain boundaries and defects, or it can be altered by quantum effects. Bulk vs Nano: In bulk materials, grain boundaries and defects often impede electron flow, which may not be the case in some nanomaterials.

\vspace{1mm}
(c) Band Gap Modulation: The band gap of nanomaterials can be tuned based on their size, especially evident in semiconductor quantum dots. Bulk vs Nano: Bulk semiconductors have a fixed band gap, while in nanomaterials, especially quantum dots, the band gap can vary based on the size of the particle.

\vspace{2mm}
In conclusion, the properties of nanomaterials are inherently influenced by their reduced dimensions, leading to a set of characteristics that are distinct from their bulk counterparts. These unique attributes are central to the expanding applications of nanomaterials in MEMS and other technological fields, unlocking functionalities and performances that were previously unattainable with bulk materials.\\

\midrule

\textcolor{black}{Surface Modification: Describe strategies used to modify the surface properties of nanomaterials in this category, including techniques like functionalization, coating, or doping. Explain how these modifications enhance their performance or enable specific applications.} &  Surface modification plays a pivotal role in tuning the properties of nanomaterials, given the significant fraction of atoms or molecules that reside on their surfaces. By tweaking the surface properties, the performance, compatibility, or functionality of nanomaterials can be optimized. Here are some common strategies:

\vspace{2mm}
(1) Functionalization: (a) Description: Functionalization involves the attachment of specific functional groups, molecules, or polymers to the surface of nanomaterials. This can be achieved through covalent bonding, electrostatic interactions, or other chemical or physical methods, (b) Applications and Performance Enhancement: Biocompatibility: Functionalization with specific organic molecules can make nanomaterials more compatible with biological systems, facilitating their use in medical MEMS devices. (c) Improved Solubility: Functionalizing nanoparticles with hydrophilic groups can enhance their dispersibility in aqueous solutions. (d) Sensing Applications: By attaching specific molecules to the surface, nanoparticles can be tailored to detect certain chemicals or biological molecules selectively.

\vspace{2mm}
(2) Coating: (a) Description: This involves enclosing or covering the nanomaterial with another material. Common coatings include polymers, metals, oxides, or other nanoparticles.
(b) Applications and Performance Enhancement: (i) Protection: Coatings can protect sensitive nanomaterials from environmental factors like oxidation or moisture. (ii) Enhanced Stability: Encapsulation can prevent aggregation of nanoparticles, ensuring consistent behavior and performance. (iii) Tailored Optical Properties: Coating can modify the optical properties, like fluorescence, of nanoparticles. This is beneficial for imaging or display applications.

\vspace{2mm}
(3) Doping: (a) Description: Doping introduces impurity atoms into the nanomaterial, deliberately modifying its electronic properties. For example, introducing phosphorus or boron atoms into silicon can change its conductivity type. (b) Applications and Performance Enhancement: (i) Modulation of Electronic Properties: Doping can convert semiconductors from being p-type (positive) to n-type (negative) or vice versa, essential for creating electronic components like transistors. (ii) Enhanced Sensing Capabilities: Doping can make nanomaterials more responsive to certain chemicals, enhancing their sensitivity in sensing applications. (iii) Improved Catalytic Activity: Doped nanomaterials can exhibit enhanced or selective catalytic behaviors, useful in chemical synthesis or environmental remediation.

\vspace{2mm}
In conclusion, the surface modification strategies like functionalization, coating, and doping offer a versatile toolkit to fine-tune the properties of nanomaterials used in MEMS. By strategically altering their surface, these nanomaterials can be optimized for specific applications or to exhibit desired behaviors, expanding their utility and effectiveness in various technological domains.

\\

\midrule

\textcolor{black}{Applications: Explore the extensive range of applications wherein nanomaterials from this category find use. Discuss their potential impact on fields such as electronics, energy, medicine, and more.} &  Nanomaterials used in MEMS have spurred a revolution across a multitude of fields due to their unique properties. Let's explore some of these applications and their impact on various sectors:

\vspace{2mm}
(1) Electronics: (a) Nano-transistors: As the demand for smaller and more powerful electronic devices grows, there's a shift towards using nanomaterials like graphene and carbon nanotubes to create transistors at the nanoscale, allowing for more densely packed circuits and improved device performance. (b) Memory Storage: Nanomaterials enable the creation of high-density memory storage devices, leading to compact storage solutions with greater capacities. (c) Sensors: Nanomaterial-based sensors can detect minute quantities of chemicals or physical changes, making them suitable for a range of applications, from environmental monitoring to electronic noses.

\vspace{2mm}
(2) Energy: (a) Solar Cells: Nanomaterials like quantum dots and nanostructured semiconductors can be used to improve the efficiency of solar cells by enhancing light absorption and charge transport. (b) Batteries and Supercapacitors: Nanomaterials increase the surface area for chemical reactions in batteries, leading to devices with higher energy storage capacities and faster charge/discharge rates. (c) Fuel Cells: Nanocatalysts improve the efficiency of fuel cell reactions, paving the way for cleaner energy sources.

\vspace{2mm}
(3) Medicine: (a) Drug Delivery: Nanoparticles can be engineered to target specific cells or tissues in the body, allowing for targeted drug delivery, which can reduce side effects and improve therapeutic outcomes. (b) Imaging: Nanoparticles with unique optical properties, such as gold nanoparticles or quantum dots, can be used for enhanced imaging techniques, aiding in the diagnosis and monitoring of diseases. (c) Biosensors: Nanoscale sensors can detect minute concentrations of biomolecules, enabling early disease detection and monitoring.

\vspace{2mm}
(4) Environmental Applications: (a) Water Purification: Nanomaterials like carbon nanotubes and nanostructured membranes can effectively filter out contaminants from water, addressing global clean water challenges. (b) Air Purification: Nanocatalysts can break down harmful pollutants in the air, contributing to cleaner urban environments. (c) Environmental Monitoring: Nanosensors can detect trace amounts of environmental pollutants, aiding in monitoring and mitigation efforts.

\vspace{2mm}
(5) Photonics and Optoelectronics: (a) Displays: Quantum dots and other nanoscale phosphors can be used in display technologies, offering brighter and more energy-efficient screens.
(b) Lasers: Nanomaterials can be used to produce lasers with specific wavelengths, useful in communications and medical procedures. (c) Photodetectors: Nanoscale detectors can respond to a broader range of light wavelengths with higher sensitivities, improving optical communication systems.

\vspace{2mm}
In conclusion, the nanomaterials in MEMS have an extensive range of applications that touch almost every aspect of modern life. From enhancing the performance of electronic devices to pioneering medical breakthroughs and addressing environmental challenges, these materials hold the promise of reshaping the future of technology and improving the quality of life on a global scale.

 \\

\bottomrule
\end{tabularx}

\end{document}